\documentclass[5p]{elsarticle}

\usepackage{lineno,hyperref}
\modulolinenumbers[5]

\journal{Journal}

\usepackage{multirow,graphicx}
\usepackage{amsmath,amssymb} % define this before the line numbering.
\usepackage{color}
\usepackage{subfigure}
\usepackage{bm}
\usepackage{breakcites}
\usepackage{url}

\begin{document}

\begin{frontmatter}

\title{A Fully Convolutional Two-Stream Fusion Network for Interactive Image Segmentation}





\author[lboro]{Yang Hu}
\ead{y.hu@lboro.ac.uk}
\author[lboro]{Andrea Soltoggio}
\ead{a.soltoggio@lboro.ac.uk}
\author[lboro]{Russell Lock}
\ead{r.lock@lboro.ac.uk}
\author[ICE]{Steve Carter}
\ead{stevec@theiceagency.co.uk}

\address[lboro]{Loughborough University, UK}
\address[ICE]{The ICE Agency, UK}

\begin{abstract}
In this paper, we propose a novel fully convolutional two-stream fusion network (FCTSFN) for interactive
image segmentation. The proposed network includes two sub-networks: a two-stream late fusion network (TSLFN) that predicts the foreground at a reduced resolution,
and a multi-scale refining network (MSRN) that refines the foreground at full resolution. The TSLFN includes two distinct deep streams followed by a fusion network.
The intuition is that, since user interactions are more direct information on foreground/background than the image itself, the two-stream structure of the TSLFN reduces the number of layers between the pure user interaction features and the network output, allowing the user interactions to have a more direct impact on the segmentation result. The MSRN fuses the features from different layers of TSLFN with different scales, in order to seek the local to global information on the foreground to refine the segmentation result at full resolution. We conduct comprehensive
experiments on four benchmark datasets. The results show that the proposed network achieves competitive performance compared to current state-of-the-art interactive image segmentation methods\footnotemark.
\end{abstract}


\end{frontmatter}

%\linenumbers

\footnotetext{The codes will be available at \url{https://github.com/cyh4/FCTSFN}.}

\section{Introduction}
\label{sec:introduction}

Binary image segmentation aims to separate an image into an object of interest (foreground) and the other parts (background).
It has a wide range of applications,~\textit{e.g.}, medical image analysis, image editing, object retrieval,~\textit{etc}.
However, since the object of interest varies highly in different contexts, most fully automatic methods are tailored and optimized to seek
the particular object of interest in a certain application. It is difficult to develop a fully automatic
method which is guaranteed to work in general applications.

To improve the flexibility and generality of image segmentation methods, many algorithms adopt interactive frameworks.
These algorithms allow users to interact with a system to specify the object of interest by labeling some foreground/background pixels.
Most traditional algorithms of interactive image segmentation~\cite{GraphCut_Boykov_2001,GrabCut_Rother_2004,GeodesicMatting_Bai_2007,GeodesicGraphCut_Price_2010,GeodesicStar_Gulshan_2010,GrowCut_Vezhnevets_2005,RandomWalk_Grady_2006}
rely on low-level features to estimate the foreground/background
distributions from user-labeled pixels, to predict the category of unlabeled pixels. A problem relating to these methods is that low level features may not be effective to distinguish between foreground and background in many situations,
~\textit{e.g.}, the foreground and the background have similar color and texture; or the foreground includes several parts with very
different appearance. Consequently, low-level feature-based algorithms may require a large number of user interactions to obtain reliable segmentation, increasing the burden on the end user.

Recently, deep features produced by deep neural networks (DNNs) have shown their power in many computer vision tasks including image classification~\cite{AlexNet_Krizhevsky_2012,VGG_Simonyan_2015,ResidueNet_He_2016}
and semantic segmentation~\cite{FCN_Shelhamer_2017,CRF_RNN_Zheng_2015,CRF_SemanticSeg_Chen_2017,InstanceAware_SemanticSeg_Li_2017}. Thus, several researchers~\cite{FCN_InteractiveSeg_Xu_2016,DeepIGeoS_Wang_2017,DeepSegmentationCaptioning_Boroujerdi_2017,DeepGrabCut_Xu_2017,DEXTR_Maninis_2018,RISNet_Liew_2017,ITIS_Mahadevan_2018} have
used DNNs to extract deep features with higher-level understanding for image and user interactions to improve interactive image segmentation.
Most of these DNN-based methods can be viewed as an early fusion of features using DNN.
They concatenate features from image and user interaction as the input to DNN; generally, a DNN is used to combine the concatenated features to predict the foreground and background. However, such early-fusion schemes may not fully exploit the information in user interactions to predict foreground/background. Specifically, considering that state-of-the-art DNNs usually consist of a large number of layers, an early-fusion of user interactions with image features may weaken the influence of user interactions on the final prediction results.

In contrast to existing networks performing early fusion, we argue that better performance can be achieved with a late-fusion structure that uses two individual deep streams to learn and extract deep features from the image and the user interactions individually, then fusing the features from the two streams. Our intuition is that such a late fusion structure allows the user interactions to have a more direct impact on the prediction result, as it has a smaller number of layers between pure user interaction features and the prediction results. We expect this will lead to improved performance, as user interactions are more direct information on the location of foreground/background than the image itself. At the same time, deep features are still produced from the two individual deep streams, so the whole network still preserves the representative advantage of deep features. This allows the network to accurately understand image content and predict the object of interest.

\begin{figure}[t]
\centering
\includegraphics[width=0.5\textwidth]{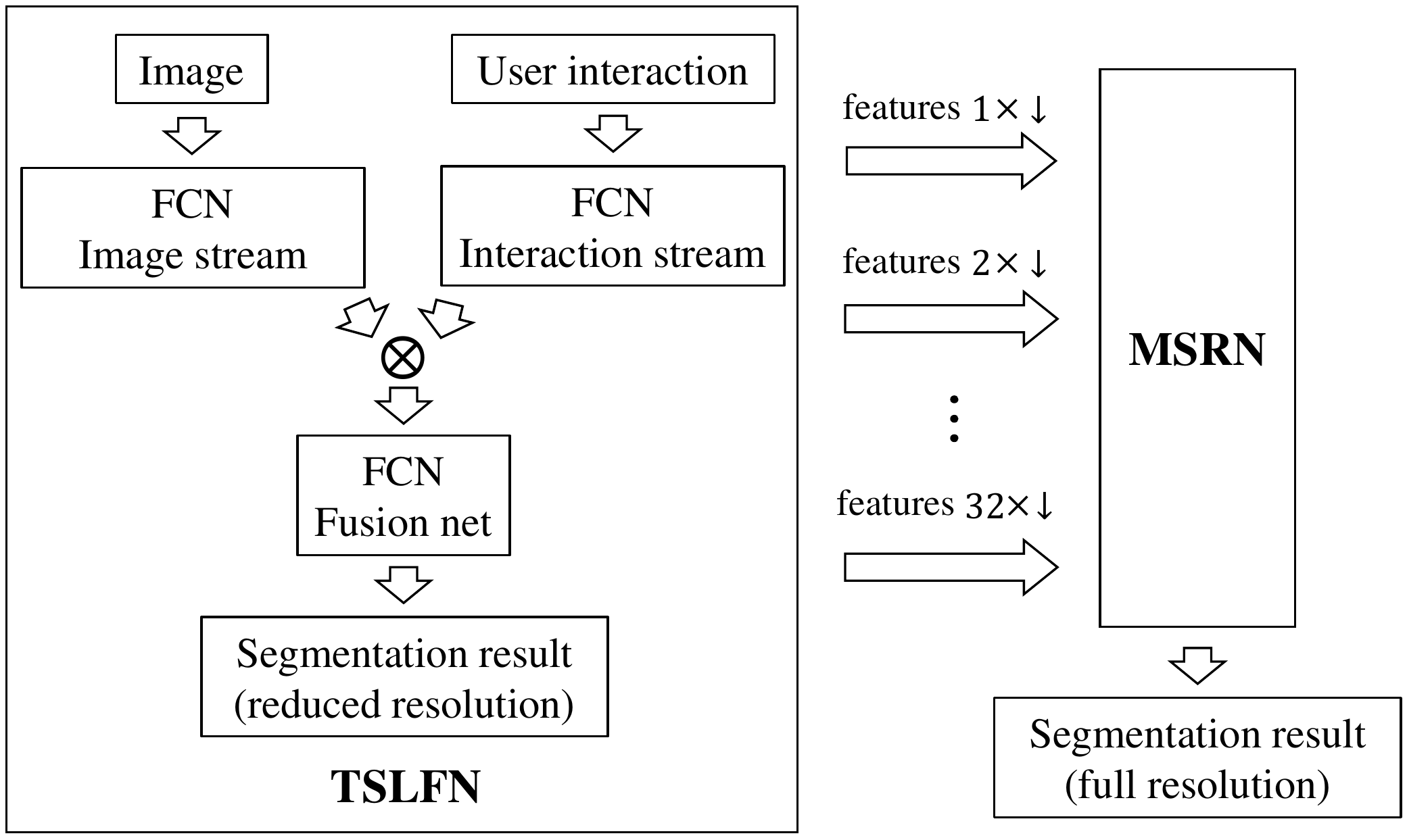}
\label{fig_network_architecture_b}
\caption{Flowchart of the proposed network architecture. $\otimes$ is the concatenation operator; $n\times\downarrow$ means $n$-time downsampling}
\label{fig_network_architecture}
\end{figure}

In this paper, we propose a novel fully convolutional two-stream fusion network (FCTSFN) for interactive image segmentation. As shown in Fig.~\ref{fig_network_architecture}, the proposed network starts with two-stream late fusion network (TSLFN). The TSLFN extracts deep features from the image and the user interaction
individually using two separate streams, and it applies a fusion net to fuse the features from the two streams to predict foreground and background. Since this two-stream late fusion structure reduces the number of layers between pure user interaction features and the network output, we expect it is able to improve the impact of user interactions on the prediction results to achieve better segmentation performance. Furthermore, to handle the loss of resolution in the TSLFN, we use a multi-scale refining network (MSRN) to refine the result of TSLFN at full resolution. The MSRN fuses the features from different layers of the TSLFN with
different scales. It is expected that the fusion result includes the local to global structural information on the object specified by user interactions, and hence the MSRN can utilize the fusion result to refine the ouput of TSLFN at full resolution. The contributions of this paper are as follows:
\begin{itemize}
    \item We propose a novel fully convolutional two-stream fusion network (FCTSFN) for interactive image segmentation.
    \item In FCTSFN, we propose a two-stream late fusion network (TSLFN) that aims to improve the impact of user interactions on the prediction results to achieve better segmentation performance.
    \item In FCTSFN, we propose a multi-scale refining network (MSRN) that fuses the information at different scales to refine the output of the TSLFN.	
\end{itemize}

The rest of the paper is organized as follows. In section~\ref{sec:related_works}, we review related works. In section~\ref{sec:proposed_method}, we detail the proposed FCTSFN for interactive image segmentation.
In section~\ref{sec:experiments}, we report the experimental results of the analysis and of the comparisons. Section~\ref{sec:conclusion} concludes the paper.

\section{Related works}
\label{sec:related_works}

A large number of methods have been proposed for interactive image segmentation. Boykov and Jolly~\cite{GraphCut_Boykov_2001} propose a graph cut-based method. This method represents the image as a graph where pixels are considered as graph nodes and neighbouring nodes are connected by edges.
With this graph structure, interactive image segmentation is formulated as an energy minimization problem which can be solved by graph cuts~\cite{GraphCut_Optimization_Boykov_2001,GraphCut_Condition_Kolmogorov_2004}.
Following~\cite{GraphCut_Boykov_2001}, Rother~\textit{et al.}~\cite{GrabCut_Rother_2004} propose GrabCut which applies graph cuts iteratively. With an initial bounding box provided by the user, GrabCut iterates between foreground/background distribution estimation and graph cut
segmentation to progressively refine foreground and background. With a similar graph representation to that in~\cite{GraphCut_Boykov_2001}, Bai and Sapiro~\cite{GeodesicMatting_Bai_2007} determine foreground and background using the geodesic distance between unlabeled pixels and user-labelled foreground/background
pixels. To take the advantage of both graph cut and geodesic distance, Price~\textit{et al.}~\cite{GeodesicGraphCut_Price_2010} propose geodesic graph cut which incorporates geodesic distance into a graph cut-based framework.
Also, Gulshan~\textit{et al.}~\cite{GeodesicStar_Gulshan_2010} improve the graph cut method~\cite{GraphCut_Boykov_2001} by applying geodesic distance with star convexity as a shape constraint for the segmentation result. In other representative research, Vezhnevets and Konouchine~\cite{GrowCut_Vezhnevets_2005} propose
GrowCut which iteratively updates labels of pixels based on cellular automaton~\cite{CA_Neumann_1966}. Grady~\cite{RandomWalk_Grady_2006} proposes a random walk method. This method calculates the probability that
a random walker firstly reaches user-labelled pixels when starting from each unlabeled pixel; the pixel labels are then assigned based on the user-labeled pixel with the highest probability. All the above methods utilize low level features (color, texture, \textit{etc.}) to
model the foreground and background distributions. Therefore, their performance is restricted by the suitability of low-level features to distinguish between the foreground and background. As a result, for complex scenes where low-level features
are less descriptive of the foreground and background differences, these methods may need users to label a large number of pixels to achieve good segmentation result. This increases the load of users.

Recently, deep neural networks (DNNs) have shown superior performance in distinguishing different objects in images~\cite{FCN_Shelhamer_2017,CRF_RNN_Zheng_2015,CRF_SemanticSeg_Chen_2017,InstanceAware_SemanticSeg_Li_2017}. Also, the deep features learned from
DNNs are proven to be highly transferable to other problems~\cite{UnderstandDNN_Zeiler_2014,FineTuning_Oquab_2014}. Hence, several researchers have focused on applying DNN to gain features with higher-level understanding for
image and user interactions to improve interactive image segmentation. Xu~\textit{et al.}~\cite{FCN_InteractiveSeg_Xu_2016} use two Euclidean distance maps to represent positive and negative clicks of the user.
They form image-user interaction pairs by concatenating the two distance maps with RGB channels of the image. A fully convolutional network (FCN) is trained to predict the foreground/background from image-user interaction pairs. With similar image-user interaction pairs as input to the network, Boroujerdi~\textit{et al.}~\cite{DeepSegmentationCaptioning_Boroujerdi_2017} use a lyncean fully convolutional network to predict foreground/background. This network replaces the last two convolutional layers in the FCN in~\cite{FCN_InteractiveSeg_Xu_2016} with three convolutional layers with gradually decreased kernel size to better
capture the geometry of objects. Wang~\textit{et al.}~\cite{DeepIGeoS_Wang_2017} transform user interactions into two geodesic distance maps. They construct image-user interaction pairs
similarly to~\cite{FCN_InteractiveSeg_Xu_2016} but augment it with an initial segmentation proposal produced by an additional DNN. They predict foreground/background using a resolution-preserving network. Xu~\textit{et al.}~\cite{DeepGrabCut_Xu_2017} propose Deep GrabCut. This method can seek the object boundary from a bounding box provided by the user. It transfers the bounding box into a distance map. An
encoder-decoder network is used to predict foreground/background from the concatenated image and distance map. Maninis~\textit{et al.}~\cite{DEXTR_Maninis_2018} seek the foreground from extreme points. They encode extreme points into 2D Gaussians which are concatenated with the input image; a residue network~\cite{ResidueNet_He_2016} and a pyramid scene parsing model~\cite{PSPNet_Zhao_2017} are used to predict the foreground. Li~\textit{et al.}~\cite{LatentDensityInteractive_Li_2018} use a segmentation network to generate various potential segmentations from image and user interactions; then a selection network is applied to select the output from the potential segmentations. Mahadevan~\textit{et al.}~\cite{ITIS_Mahadevan_2018} propose an iterative training algorithm. Instead of training with fixed user clicks, this algorithm adds clicks progressively based on the error of the network predictions. This algorithm leads to improved performance, since it is more closely aligned with the patterns of real users. Essentially, all these networks~\cite{FCN_InteractiveSeg_Xu_2016,DeepSegmentationCaptioning_Boroujerdi_2017,DeepIGeoS_Wang_2017,DeepGrabCut_Xu_2017,DEXTR_Maninis_2018,ITIS_Mahadevan_2018, LatentDensityInteractive_Li_2018} adopt early fusion structures. They combine the image and the user interaction features from the first layer of DNN. Differently from them, the proposed FCTSFN extracts deep image and user interactions features individually and then fuse them.

The work most similar to the proposed network is that of Liew~\textit{et al.}~\cite{RISNet_Liew_2017}. This is also a two-branch network: it includes a global branch producing coarse global predictions and a local branch utilizing multi-scale spatial pyramid features to make refined local predictions; the final prediction is the combined results from the two branches. However, the proposed network differs from the network in~\cite{RISNet_Liew_2017} in three important aspects. First, Liew~\textit{et al.} concatenate the image and interaction maps as the input of the network; the proposed network uses two individual streams to extract features from the image and interaction maps, to allow user interactions to have a more direct impact on the segmentation results. Second, Liew~\textit{et al.} produce multi-scale features using a spatial pyramid pooling on the features at an end
layer of the network; the proposed network utilizes and fuses the features from different layers of the network, to incorporate both low-level information like color and edges and higher-level object information into the foreground prediction. Third, Liew~\textit{et al.} use multi-scale features to refine local segmentations which are then combined with global segmentation; the proposed network uses the multi-scale features in a direct global prediction refinement structure to make the prediction at full-resolution.

\section{The proposed network}
\label{sec:proposed_method}

In this section, we present the proposed fully convolutional two-stream fusion network (FCTSFN) for interactive image segmentation. Firstly, we describe the architecture of the two-stream late fusion network (TSLFN) in the overall FCTSFN architecture. Then, we present the structure of the multi-scale refining network (MSRN) in the FCTSFN.
Next, we demonstrate the network training process. Finally, we describe the data processing for the whole FCTSFN, including the method to generate user interaction maps from user interactions as input to the network and the method
to produce the foreground mask based on the output of the network.

\begin{figure*}[!t]
\centering
\subfigure[TSLFN]
{
\includegraphics[width=0.85\textwidth]{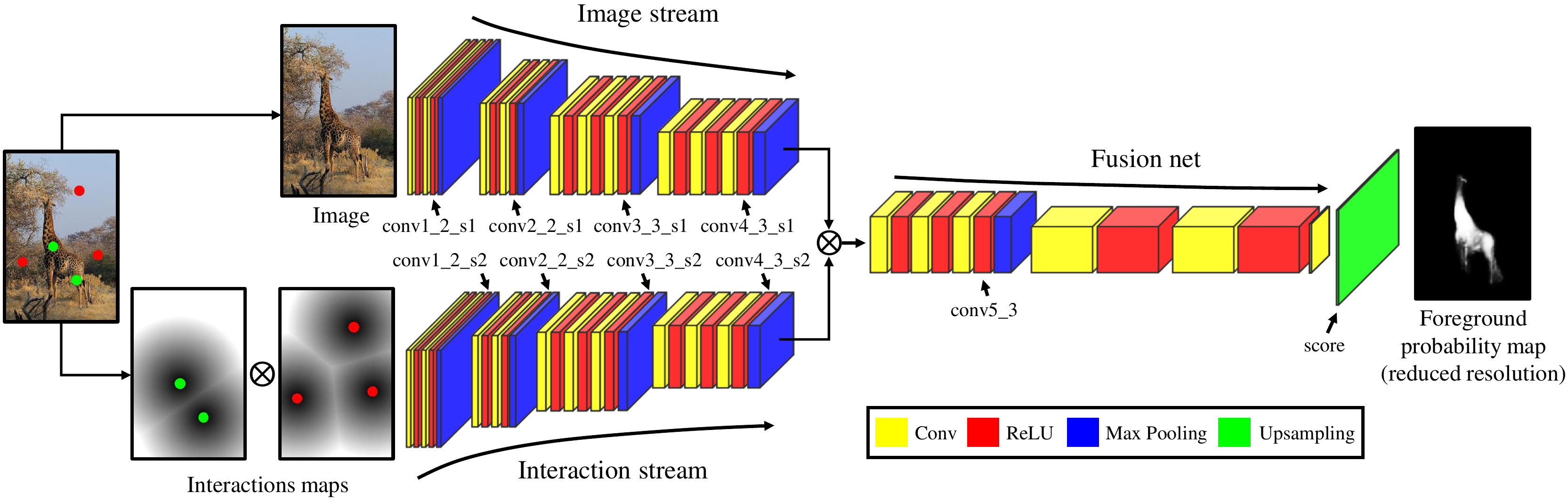}
\label{fig_TSLFN}
}
\subfigure[MSRN]
{
\includegraphics[width=0.7\textwidth]{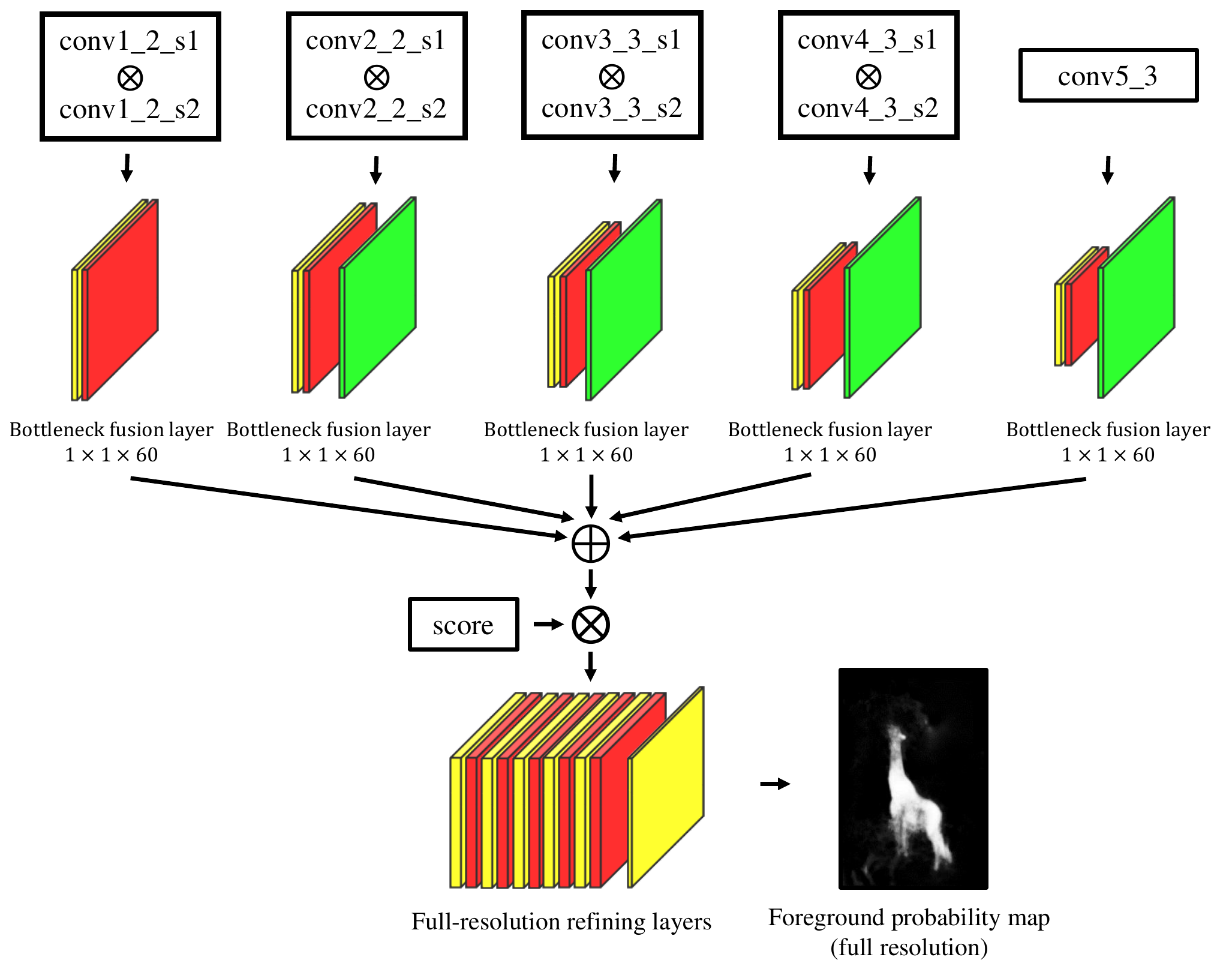}
\label{fig_MSRN}
}
\caption{Architectures of the proposed fully convolutional two-stream fusion network (FCTSFN). It includes a two-stream late fusion network (TSLFN) and a multi-scale refining network (MSRN). $\oplus$ is element-wise sum operation. The sizes of convolutional layers are in the format of filter height $\times$ filter width $\times$ number of filters}
\label{fig_FCTSFN}
\end{figure*}

\subsection{Two-stream late fusion network (TSLFN)}
\label{sec:proposed_method_TSLFN}

Fig.~\ref{fig_TSLFN} presents the structure of the TSLFN in the proposed FCTSFN. The input of TSLFN has two parts: one is an image and the other one is the concatenated positive and negative interaction maps generated respectively from positive and negative user interactions (see section~\ref{sec:proposed_method_data_processing}).
The network outputs a foreground probability map at a reduced resolution indicating the likelihood that a pixel is foreground. This network uses the VGG16 network~\cite{VGG_Simonyan_2015} as the base network. It includes three parts: an image stream, an interaction stream, and a fusion net. The network in either image or interaction stream consists of the first 10
convolutional layers of the VGG16 network with rectified linear unit (ReLU) activation.
After several convolutional layers, a max-pooling layer is applied with a kernel size of $2 \times 2$ and a stride of 2. The intent of the two streams is to learn deep features for images and interaction maps individually.

At the end of the image and the interaction streams, the feature maps from both streams are concatenated. The fusion net is then applied to learn to combine the concatenated feature maps to predict foreground/background. The fusion net consists of
6 convolutional layers: the first 3 of them are from the last 3 convolutional layers of the VGG16 network (corresponding to \texttt{\small{conv5\_1}}, \texttt{\small{conv5\_2}}, \texttt{\small{conv5\_3}} in the VGG16);
the last 3 of them are transferred from the fully connected layers in the VGG16 network using the method in~\cite{FCN_Shelhamer_2017}. Since the
output of the whole network is downsampled by a factor of $32$ with respect to the input image, we use an upsampling layer to upscale the output to the original resolution. Similarly to~\cite{FCN_Shelhamer_2017}, the upsampling layer is a
deconvolution layer with the filters set to bilinear interpolation kernels.

Note that variations of this TSLFN structure can be devised. One can make different assignments of the layers of the VGG16 network between the image/interaction stream and the fusion net to create variations of TSLFN with different depth in the two streams and the fusion net. This is essentially a trade-off between the impact of user interactions and the prediction capacity of the network. If we use fewer layers in the fusion net, the location information in user interactions may have higher impact on the prediction results, since it reduces the number of layers between pure user interaction features and the prediction result. However, this may harm the prediction capacity of the network, since the fusion net is too shallow and it may not be able to learn an effective projection from image/user interaction features to the foreground. On the other hand, a deeper fusion net may have sufficient capacity to learn a projection from image/user interaction features to the foreground, but it may weaken the influence of the user interactions on the prediction result due to the increasing number of layers between the pure user interaction features and the prediction result. Experimentally, we find that the structure shown in Fig.~\ref{fig_TSLFN} provides top performance compared to its other variations (see section~\ref{sec:experiments_analysis_TSLFN}). We assume that this is because it achieves the best trade-off between the impact of user interactions and the prediction capacity, given our base network.

Also, we note that another possible way to fuse the image and user interaction features is the layer-by-layer fusion proposed by Hazirbas~\textit{et al.}~\cite{FuseNet_Hazirbas_2016}. This method uses two individual streams and fuses the features from the two streams multiple times at different layers. We have conducted experiments with this fusion architecture in early stage of our experiments (\textit{i.e.}~we perform similar layer-by-layer fusion between the image and user interaction streams with our base network). We found that it led to a performance drop for the task of interactive image segmentation. Considering that this architecture is originally designed to handle RGB-depth (RGBD) data, we think it is the differences in the characteristics of data that lead to the performance drop. For the RGBD data, the depth data includes accurate object boundary information, hence fusing it layer-by-layer with images data enhances the object boundary information, as discussed in~\cite{FuseNet_Hazirbas_2016}. However, for our task of interactive segmentation, the interaction maps do not include such accurate object boundary information; as a result, it is possible that fusing the features in the interaction stream into the image stream layer-by-layer brings difficulties for the network to learn information about objects.

\subsection{Multi-scale refining network (MSRN)}

The MSRN in the proposed FCTSFN aims to fuse the information at different scales in the TSLFN in order to gain a better understanding of the location of the foreground and refine the predicted foreground at full resolution. Fig.~\ref{fig_MSRN} presents the architecture of the MSRN. The MSRN uses the features in six different scales from the TSLFN: the concatenated feature maps before each pooling layer of the image and the interaction streams (see \texttt{\small{conv1\_2\_s1}}, \texttt{\small{conv1\_2\_s2}}, \texttt{\small{conv2\_2\_s1}}, \texttt{\small{conv2\_2\_s2}}, \texttt{\small{conv3\_3\_s1}}, \texttt{\small{conv3\_3\_s2}}, \texttt{\small{conv4\_3\_s1}}, \texttt{\small{conv4\_3\_s2}} in Fig.~\ref{fig_FCTSFN}); the feature maps before the pooling layer in the fusion net (\texttt{\small{conv5\_3}} in Fig.~\ref{fig_FCTSFN}); the upscaled prediction scores (\texttt{\small{score}} in Fig.~\ref{fig_FCTSFN}). The features at each scale, except the predictions scores, are passed through a convolutional layer with a size of $1\times1\times60$ (filter height $\times$ filter width $\times$ number of filters), and the feature maps with downsampling are upscaled to the original resolution (the second row in Fig.~\ref{fig_MSRN}). We call these layers ``bottleneck fusion layers'' due to their two-fold effects. First, they fuse the feature maps from the image and the interaction streams to seek the information of the foreground on a specific scale. Second, they act as bottleneck layers to reduce the dimension of feature maps to keep the computational cost feasible.

After the bottleneck fusion layers, the feature maps from different scales are fused by an element-wise sum operation. Then, the fused features are concatenated with the prediction scores from the TSLFN. Finally, the concatenated features are passed through the full-resolution refining layers to predict the refined foreground (the last row in Fig.~\ref{fig_MSRN}). The full-resolution refining layers consist of a stack of six convolutional layers. The size of these convolutional layers are $7\times7\times64$, $5\times5\times64$, $3\times3\times64$, $3\times3\times64$, $3\times3\times64$, $1\times1\times2$. We use filters with large to small size to capture information from coarse to fine regions. The last convolutional layer works as a classifier.

\subsection{Network training}

We train the FCTSFN in two stages. In the first stage, we remove the MSRN and fine-tune the TSLFN from the pre-trained VGG16 base network. In the second stage, we fix the parameters in the TSLFN and train the MSRN from scratch.

{\bf Fine-tuning TSLFN}. We fine-tune the TSLFN (Fig.~\ref{fig_TSLFN}) from the VGG16 network pre-trained on the ImageNet dataset~\cite{VGG_Simonyan_2015,ImageNet_Russakovsky_2015}.
We use pixel-wise softmax loss. We adopt the ``heavy'' learning scheme in~\cite{FCN_Shelhamer_2017} using a batch size of $1$ and a momentum of $0.99$, due to the reported effectiveness of this method in fine-tuning FCN for image segmentation tasks~\cite{FCN_Shelhamer_2017}. At this stage, due to the differences in shapes, the pre-trained weights are not directly applicable to the following layers:
the first convolutional layer in the interaction stream, the first convolutional layer in the fusion net, and the last convolutional layer in the fusion net. For the first convolutional layer in the interaction stream with a two-channel input, we use the mean of the filters in the first convolutional layer of the pre-trained VGG16 network to initialize it. The first convolutional layer in the fusion net has a doubled number of channels compared to the corresponding layer in VGG16 (\texttt{\small{conv5\_1}}), due to the concatenation of feature maps from the two streams. To initialize this layer, we divide the channels of this layer into two halves and copy the pre-trained weights in \texttt{\small{conv5\_1}} layer of VGG16 to each half. For the last convolutional layer in the fusion net, we initialize it with all zeros. Furthermore, we employ similar methods in~\cite{FCN_Shelhamer_2017} to fine-tune a stride-16 network and a stride-8 network for TSLFN to incorporate the features with finer scales to predict the foreground. We use the stride-8 network of TSLFN as the final form of TSLFN to make predictions with MSRN.

{\bf Training MSRN from scratch}. With TSLFN parameters trained and fixed, we then train MSRN from scratch. We find that, at this stage, class-imbalance has a significant impact on the training performance. Specifically, the foreground usually occupies a relatively smaller region than the background in the task of interactive image segmentation. This leads to far more background pixels than the foreground pixels in our training data (see section.~\ref{sec:experiments_setting}). Consequently, the learned network is easily biased to the background, resulting predictions with all pixels being background without foreground. To account for this problem, we crop the image centered at the foreground, if the area of the bounding box of the foreground occupies less than $35\%$ of the image area. To avoid overfitting, we use three further strategies in the training. (1) Data augmentation: before a forward-backward pass, the training images used in this pass have a $50\%$ probability to receive a random rotation, and a $50\%$ probability to receive a random translation. (2) Dropout: each convolutional layer belonging to the full-resolution refining layers in the MSRN (see Fig.~\ref{fig_MSRN}) are followed by a dropout layer with a dropout ratio of $0.5$, except the last layer for classification. (3) Early stopping: we record the validation accuracy on the validation data every 1000 iterations, and we terminate the training when we observe no improvement on accuracy in several consecutive validations. The other settings for the training of MSRN are as follows. We initialize all the convolutional layers randomly using the method in~\cite{MSAR_init_Zheng_2015}. We resize all the training images to a resolution of $240\times320$ (height$\times$width), and we train with a batch size of 3. We set the initial learning rate to $1\mathrm{e}{-8}$, weight decay to $0.0005$, and momentum to $0.99$.

\subsection{Data processing}
\label{sec:proposed_method_data_processing}

The data processing in this paper includes two parts: the generation of user interaction maps and the post-processing of the network output. We employ the method in~\cite{FCN_InteractiveSeg_Xu_2016} to generate interaction maps from user clicks as input to the network.
Given an image and user clicks, the sets of positive and negative clicks are transferred into a positive and a negative interaction map, respectively. Either interaction map has the same height and width as the input image.
Let ${\mathcal{S}}$ be a set of either positive or negative clicks. Let $s_{ij} \in \mathcal{S}$ be a click in ${\mathcal{S}}$ at coordinate $\left ( i,j \right )$.
Let $Y_{m,n}$ be the element at location $\left ( m,n \right )$ in the matrix of the interaction map corresponding to the image and the clicks in ${\mathcal{S}}$. $Y_{m,n}$ is calculated by:
\begin{equation}
Y_{m,n} = \mathop { \min }\limits_{s_{i,j} \in \mathcal{S}} \sqrt {\left ( m-i \right )^2 + \left ( n-j \right )^2}
\label{eqn:1}
\end{equation}
In other words, the interaction maps are calculated using the minimum Euclidean distance between pixels and the user clicks. The pixel values in positive and negative
interaction maps are truncated to 255. If no negative clicks are received, all pixel values in the negative interaction map are set to 255. Examples of positive and negative interaction maps are included in Fig.~\ref{fig_TSLFN}. For the post-processing of the network output, we adopt a graph cut-based method similar to the one in~\cite{FCN_InteractiveSeg_Xu_2016}.

\section{Experiments}
\label{sec:experiments}

In this section, we show the experimental analysis and comparisons for the proposed method. Firstly, we describe the experimental settings. Then, we perform experimental analysis for the proposed TSLFN and MSRN. Finally, we make comparisons to state-of-the-art algorithms for interactive image segmentation.

\subsection{Experimental settings}
\label{sec:experiments_setting}

{\bf Datasets}. We conduct experiments on four datasets: Pascal VOC 2012~\cite{VOC_Dataset_Kolmogorov_2004}, Microsoft Coco~\cite{MicrosoftCoCo_2014}, Grabcut~\cite{GrabCut_Rother_2004} and Berkeley~\cite{BerkeleyInteractive_Dataset_Kolmogorov_2004}. Pascal VOC 2012 and Microsoft Coco are benchmark datasets for object segmentation. Grabcut and Berkeley are benchmark datasets for interactive image segmentation. For Pascal VOC 2012, we employ its training set with 1464 images and validation set with 1449 images; for Microsoft Coco, we randomly select 20 images from each of its 80 categories, similarly to the setting in~\cite{FCN_InteractiveSeg_Xu_2016}; for Grabcut with 50 images and Berkeley with 100 images, we use all the images.

{\bf Data partition}. We partition the data in the above datasets into training/validation/test data as follows. We use the training set of Pascal VOC 2012 as training/validation data. From the 1464 images, we randomly select $200$ images as our validation data, and the rest images are used as our training data. We use the training data
to train neural networks. We use the validation data to monitor and control the training process. Since it is practically too expensive to collect interaction data from real users for network training, we employ the method in~\cite{FCN_InteractiveSeg_Xu_2016} to generate synthetic user interactions for the objects in the training/validation data. We use the data apart from the training/validation data as the test data for performance evaluation (\textit{i.e.}~Pascal VOC 2012 validation set, Microsoft Coco, Grabcut, Berkeley).

{\bf Performance evaluation}. We use intersection of union (IoU) of foreground to measure segmentation accuracy. It calculates the ratio of intersection between segmentation result and ground truth mask to the union of them. Based on IoU, we evaluate the performance of algorithms using two measures: foreground IoU vs.~number of clicks, and number of clicks to achieve a certain foreground IoU. The former measure demonstrates the segmentation accuracy with respect to the number of user clicks; the latter measure shows the amount of user effort to achieve a certain segmentation accuracy. Given an object in an image, we automatically generate a sequence of clicks as user interactions (see below). We track foreground IoU vs.~number of clicks, and we record the number of clicks to achieve a certain foreground IoU. For a dataset, we calculate the mean of each measure over all objects in this dataset. In this paper, we set 20 as the maximum number of clicks. For the second measure, if the certain IoU cannot be achieved in 20 clicks, we threshold the recorded number of clicks by 20.

{\bf Generating click sequences}. We designed a method to automatically generate a sequence of clicks for a given object for performance evaluation. This method iterates between (1) adding a click into the click sequence based on the current segmentation result and (2) renewing the segmentation result using the updated click sequence.
Given the current segmentation mask and the ground truth mask, we add a click as follows. Firstly, we find the false positive and the false negative regions in
the current segmentation result. Then, we select the largest connected component among the false positive and the false negative regions. We place the click at the point which is within the selected region and farthest from the boundary of this region. This click is set to a positive click if it is placed at a false
negative region, otherwise it is set to a negative click. After the click is added, we use the algorithm in evaluation to update the segmentation mask. The intuition of this method is that the added click focuses on the largest error region and it is placed in the central part of the region as much as possible.

\subsection{Analysis of TSLFN}
\label{sec:experiments_analysis_TSLFN}

Recall that our intuition to use a late fusion structure in TSLFN is to improve the impact of user interactions on the prediction result, as user interactions are more accurate information on the location of foreground/background. Therefore, as the first experiment in this subsection, we compare the impact of user clicks on the prediction result between two-stream and single-stream networks. The idea to measure the impact of user clicks on the prediction result is straightforward: a positive click has a higher impact if its surrounding region has higher response in the foreground probability map produced by the network; a negative click has a higher impact if its surrounding region has lower response in the foreground probability map. Therefore, if both positive and negative clicks achieve high impact on the network output, the responses around the two type of clicks in the foreground probability map should have well-separated distributions. Accordingly, the overall influence of positive and negative clicks can be measured by: (1) calculating the distributions of the responses in the regions around positive and negative clicks in the foreground probability map; (2) measuring how well the distributions corresponding to positive and negative clicks are separated. In this paper, we adopt decidability index (DI) to measure the degree of separation between the two distributions~\cite{IrisRecognition_Daugman_2004}:
\begin{equation}
\text{DI}=\frac{{\left | {\mu}_{p}-{\mu}_{n} \right |}}{\sqrt{\frac{{\sigma}_{p}+{\sigma}_{n}}{2}}}
\label{eqn:2}
\end{equation}
where ${\mu}_{p}$ and ${\mu}_{n}$ are the mean of the response distribution around positive and negative clicks, respectively; ${\sigma}_{p}$ and ${\sigma}_{n}$ denote the variance of the two distributions.

\begin{figure*}[!t]
\centering
\subfigure[Pascal VOC 2012]
{
\includegraphics[width=0.24\textwidth]{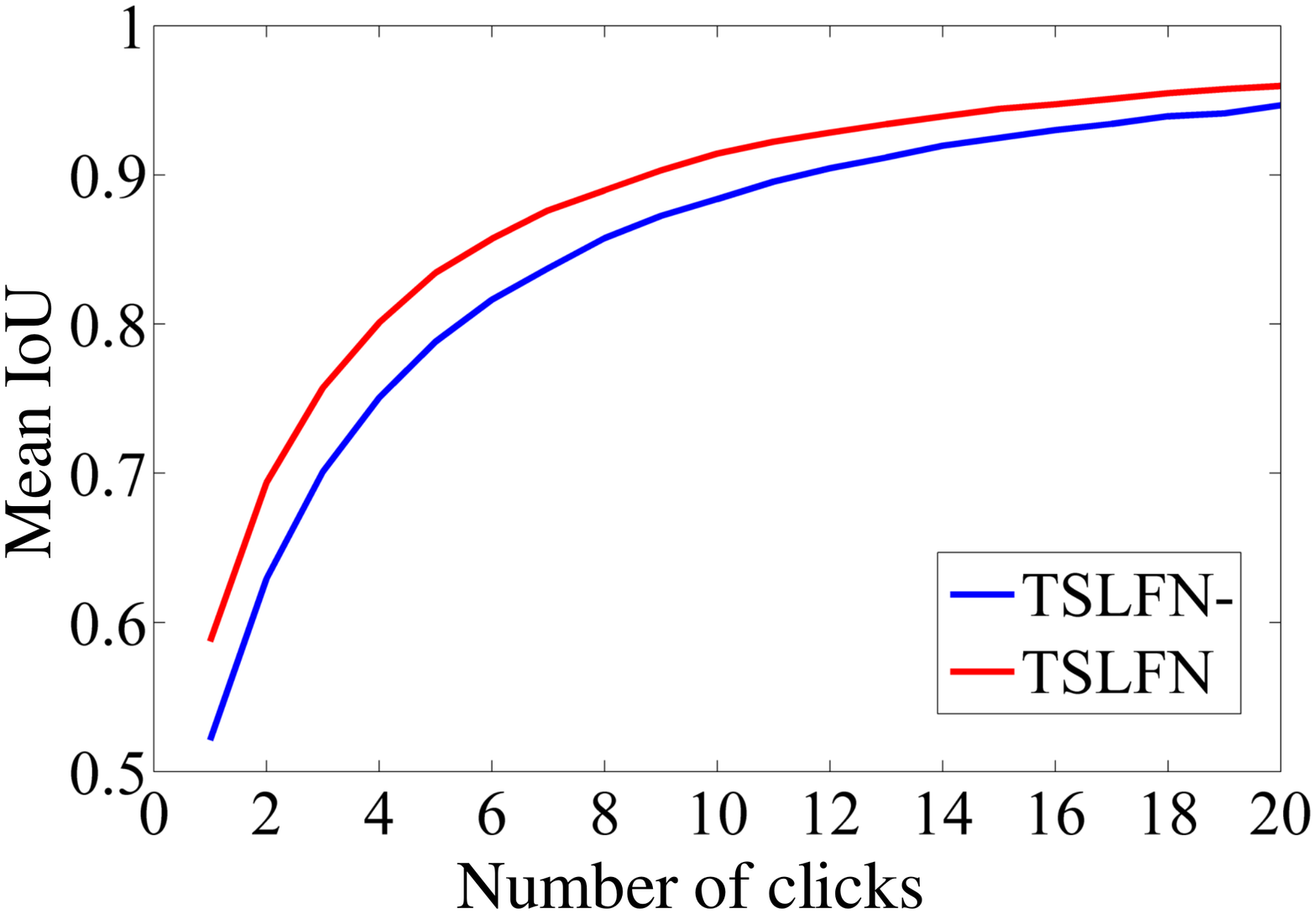}
}
\hspace{-4mm}
\subfigure[Microsoft Coco]
{
\includegraphics[width=0.24\textwidth]{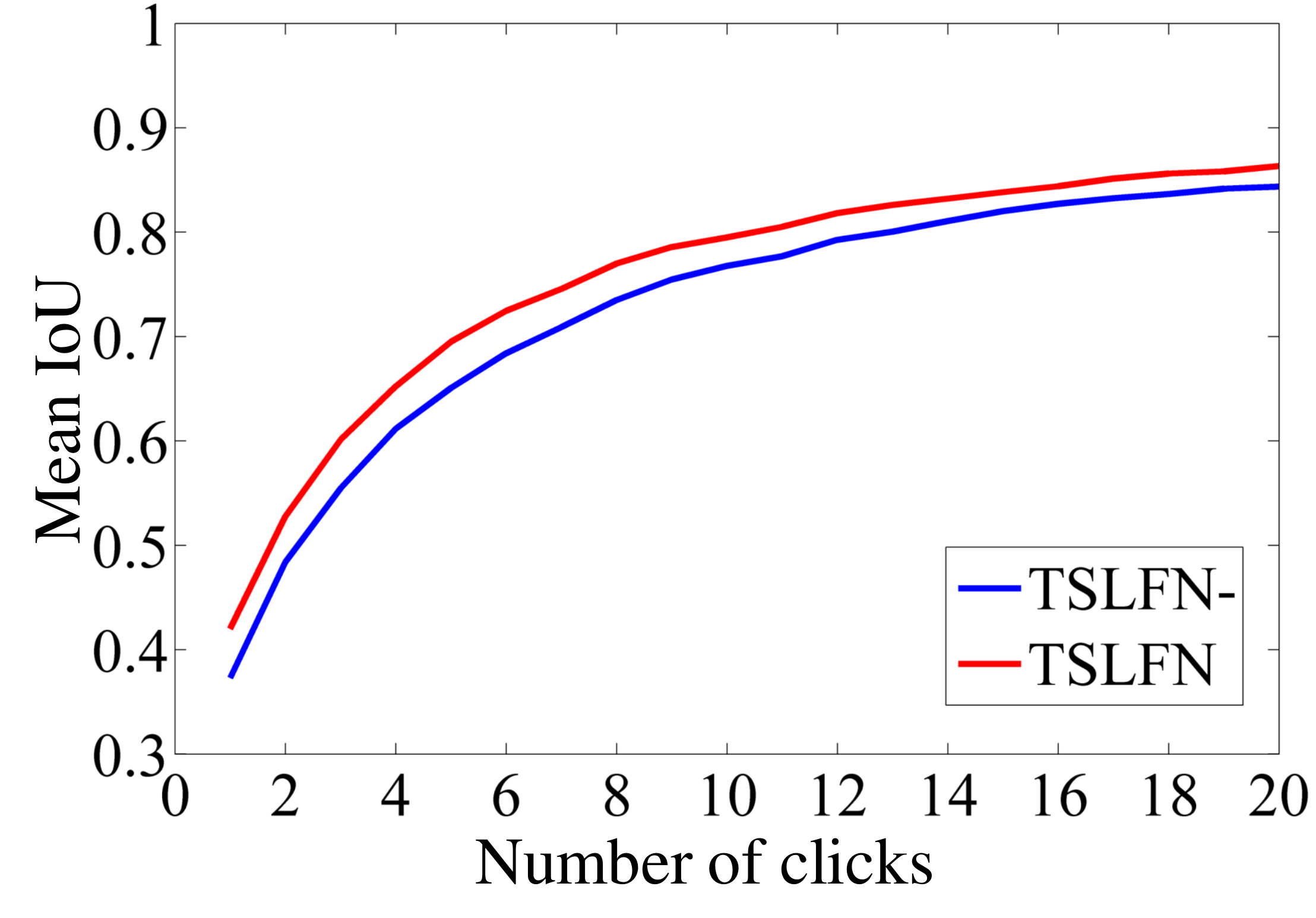}
}
\hspace{-4mm}
\subfigure[Grabcut]
{
\includegraphics[width=0.24\textwidth]{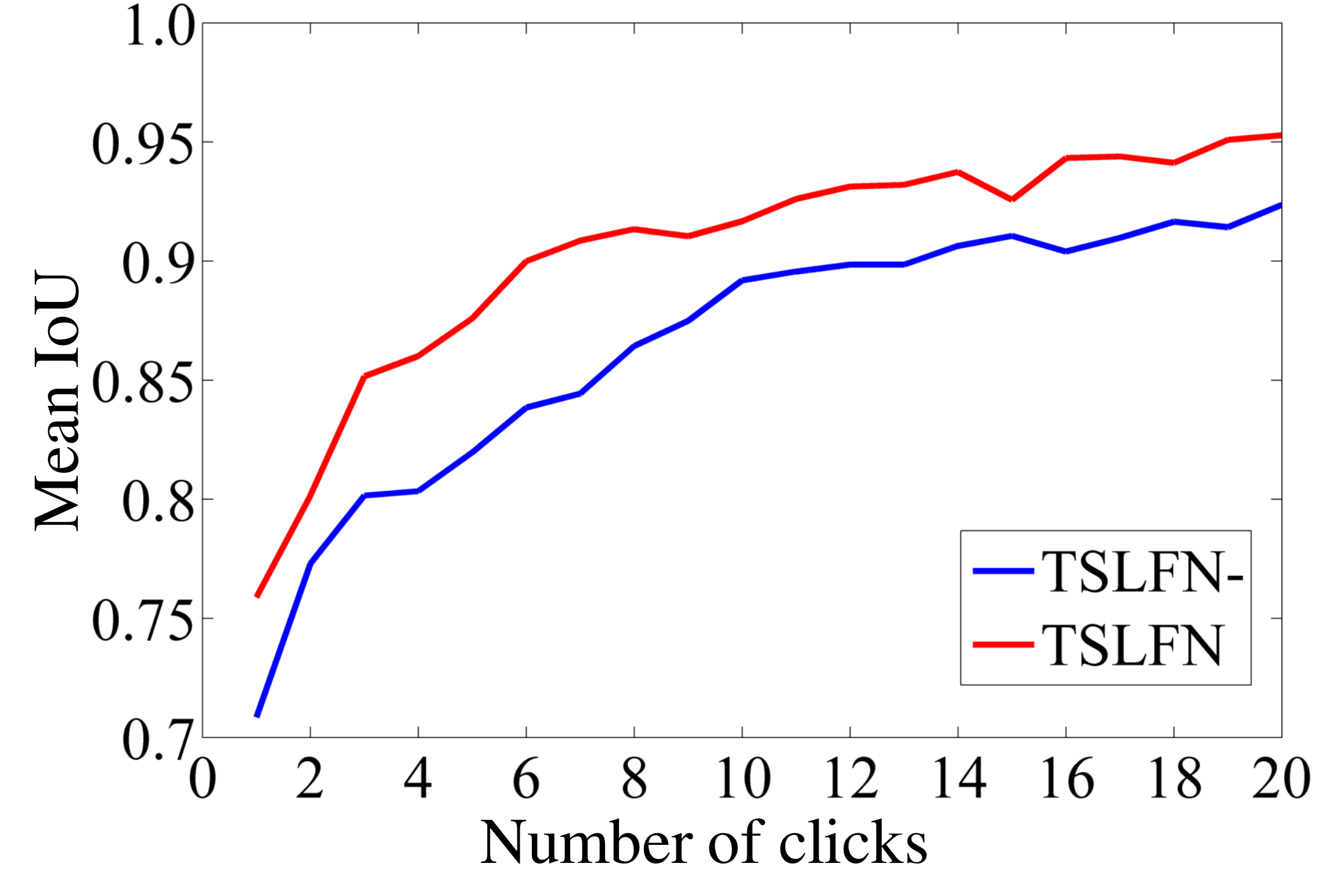}
}
\hspace{-4mm}
\subfigure[Berkeley]
{
\includegraphics[width=0.24\textwidth]{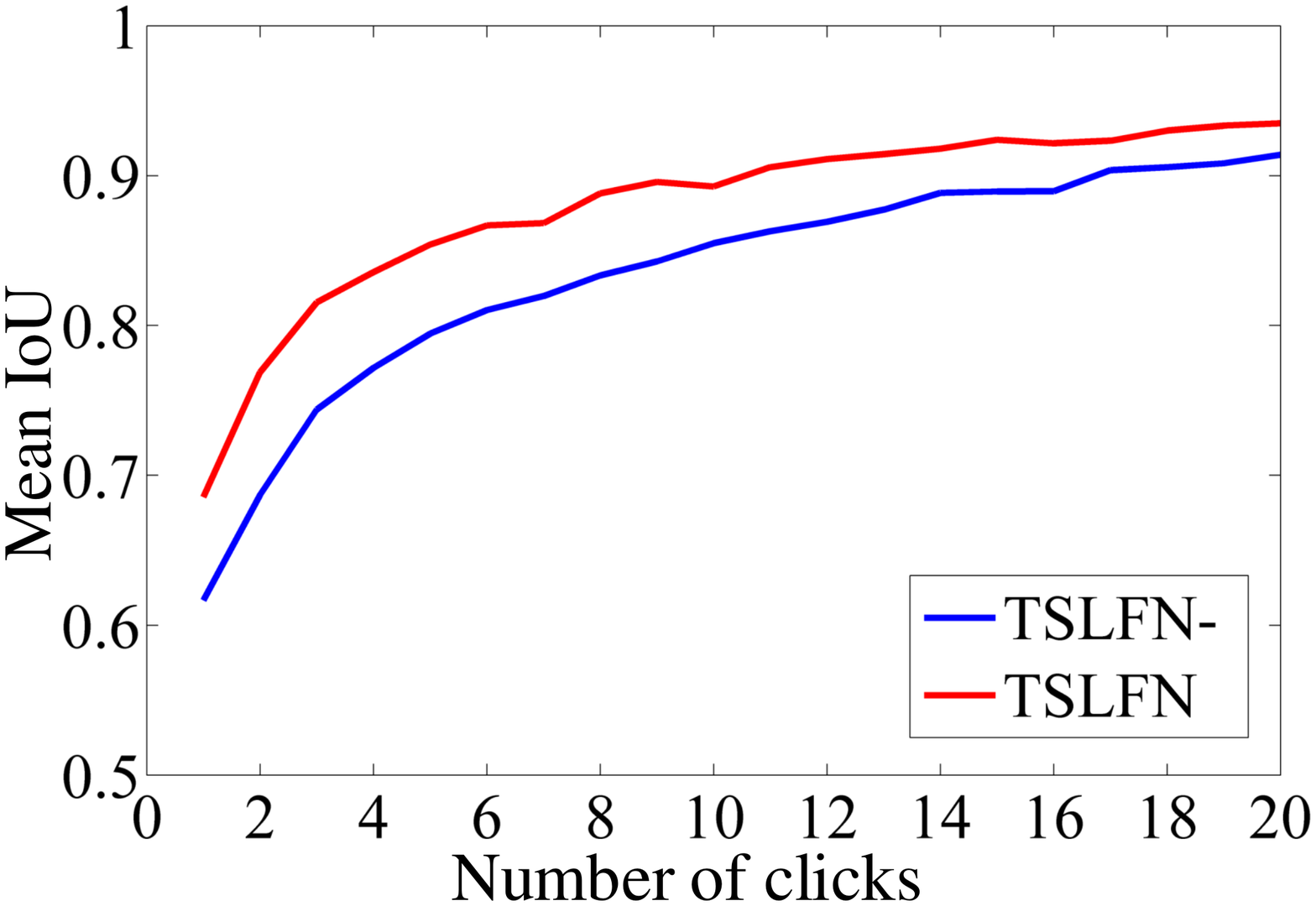}
}
\caption{Mean IoU vs.~number of clicks for the analysis of the effect of deep features from the interaction stream of the TSLFN}
\label{fig_experiment_ablation_analysis}
\end{figure*}

\begin{table*}[!t]
\begin{center}
\caption{Decidability index (DI) between regions around positive and negative clicks in the probability map (best performance in bold for each setting)}
\label{tab_experiment_DI_TSLFN}
\begin{tabular}{ccccccc}
\hline\noalign{\smallskip}
\multirow{2}{*}{Dataset} & \multicolumn{2}{c}{Free-choice} & \multicolumn{2}{c}{All-positive} & \multicolumn{2}{c}{Single-positive} \\
%\cline{2-7}
& SSFCN & TSLFN & SSFCN & TSLFN & SSFCN & TSLFN \\
\noalign{\smallskip}
\hline
\noalign{\smallskip}
Pascal VOC 2012 (1 click) & 12.97 & {\bf 15.60} & 12.97 & {\bf 15.60} & - & - \\
Microsoft Coco (1 click) & 16.79 & {\bf 20.37} & 16.79 & {\bf 20.37} & - & - \\
Grabcut (1 click) & 19.38 & {\bf 21.38} & 19.38 & {\bf 21.38} & - & - \\
Berkeley (1 click) & 71.42 & {\bf 90.91} & 71.42 & {\bf 90.91} & - & - \\
\hline
Pascal VOC 2012 (5 clicks) & 2.57 & {\bf 6.40} & 2.79 & {\bf 4.03} & 2.16 & {\bf 2.92} \\
Microsoft Coco (5 clicks) & 2.72 & {\bf 9.61} & 2.40 & {\bf 3.58} & 1.29 & {\bf 2.10} \\
Grabcut (5 clicks) & 3.84 & {\bf 7.91} & 3.14 & {\bf 4.66} & 1.89 & {\bf 3.04} \\
Berkeley (5 clicks) & 2.85 & {\bf 5.22} & 3.05 & {\bf 4.33} & 1.33 & {\bf 2.23} \\
\hline
Pascal VOC 2012 (10 clicks) & 1.24 & {\bf 2.33} & 2.33 & {\bf 3.20} & 2.34 & {\bf 3.03} \\
Microsoft Coco (10 clicks) & 1.20 & {\bf 2.37} & 2.06 & {\bf 2.87} & 1.30 & {\bf 1.89} \\
Grabcut (10 clicks) & 1.53 & {\bf 3.16} & 2.15 & {\bf 3.09} & 1.94 & {\bf 3.06} \\
Berkeley (10 clicks) & 1.33 & {\bf 2.46} & 2.03 & {\bf 2.88} & 1.50 & {\bf 2.21} \\
\hline
\end{tabular}
\end{center}
\end{table*}

To analyse the impact of user clicks between two-stream and single-stream networks, we compare the DI between TSLFN and its single-stream modification. To modify the TSLFN to a single-stream network, we remove the interaction stream, and we concatenate the interaction maps with the image at the beginning of the network. Note that the TSLFN after this modification is equivalent to the single-stream fully convolutional network (SSFCN) in~\cite{FCN_InteractiveSeg_Xu_2016}. Thus, we refer to it as SSFCN. We measure the DI with 1, 5 and 10 user clicks automatically generated using the method in section~\ref{sec:experiments_setting} (we refer to this setting as free-choice). We calculate the DI based on the responses within a radius of 10 to positive and negative clicks in the probability map. If no negative clicks exist, we calculate the DI between the responses around the positive clicks and the responses in all background regions. We calculate the DI using the above methods for each object individually, and we report the mean DI on each of our test datasets.

In addition to the free-choice setting that allows the free choices of user clicks and leads to a combination of positive and negative clicks, we also study the impact of the clicks when only positive or negative clicks exist. This is to study the behaviour of the network for different types of user clicks. To study the case that only positive clicks exist, we force all the clicks to be put on the foreground (referred to as all-positive); we measure the DI between the regions around the positive clicks and the whole background region. To investigate the case that only negative clicks exist, we consider an approximate setting: we force the first click to be on the foreground and the rest clicks to be on the background (referred to as single-positive); and we measure the DI between the regions around the negative clicks and the whole foreground region. The reason to use an approximate setting is that all of our training data have at least one positive click given the method to generate them (see section~\ref{sec:experiments_setting}), hence the network is not well trained to handle the data without positive clicks.

Tab.~\ref{tab_experiment_DI_TSLFN} shows the DIs on the four datasets with the user click settings of free-choice, all-positive and single-positive. We can see that TSLFN has a consistently higher DI compared to SSFCN. This means that the distributions of the responses around positive and negative clicks are more separated in the probability maps of TSLFN. In other words, the user clicks have higher impact with the two-stream network structure. This is consistent with our intuition of using a two-stream network. Also, we find that this trend holds for all three settings of user clicks. This shows that the improvement on the impact of user clicks achieved by the two-stream network is consistent for different types of clicks.

\begin{figure*}[!t]
\centering
\subfigure[Pascal VOC 2012]
{
\includegraphics[width=0.24\textwidth]{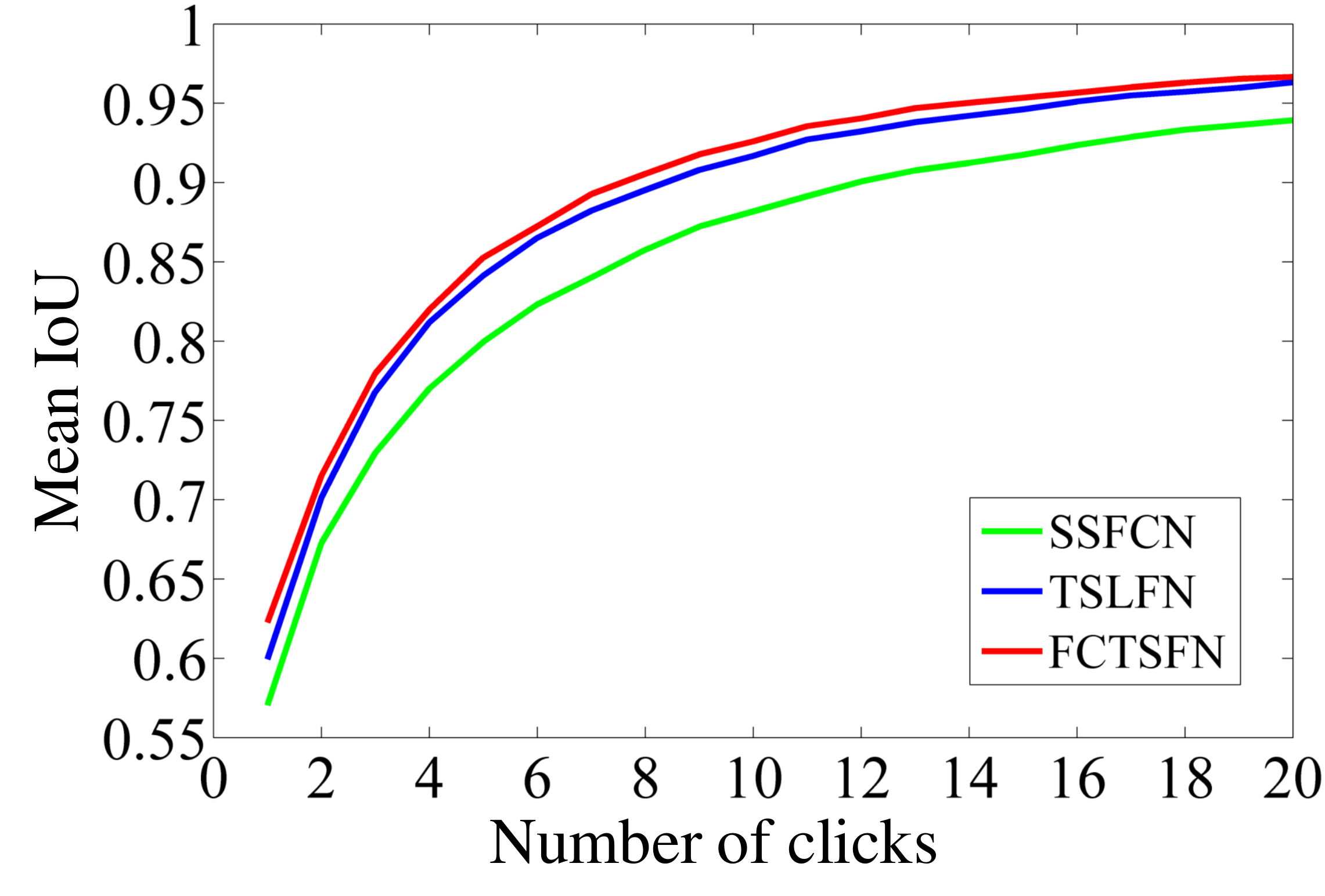}
}
\hspace{-4mm}
\subfigure[Microsoft Coco]
{
\includegraphics[width=0.24\textwidth]{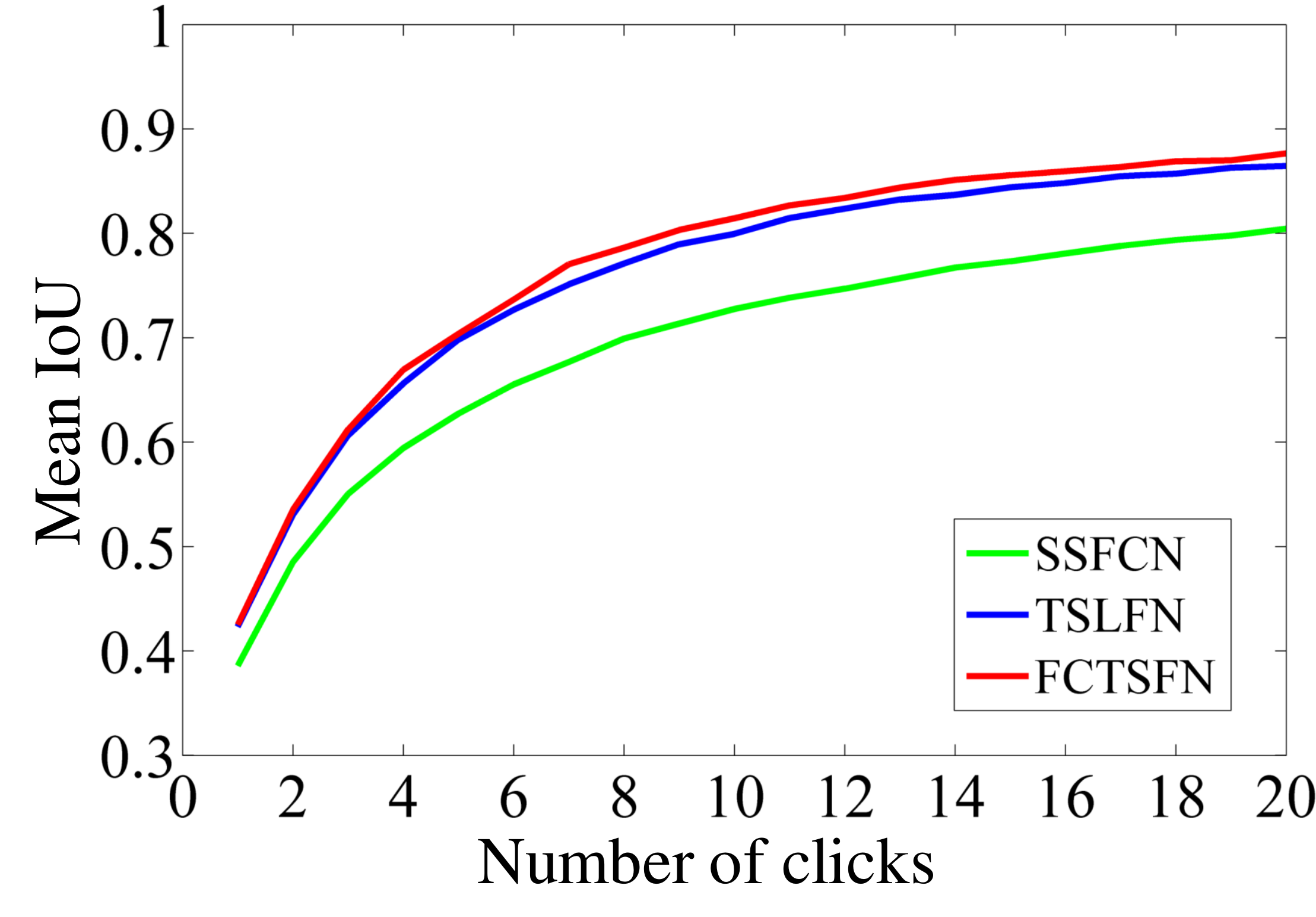}
}
\hspace{-4mm}
\subfigure[Grabcut]
{
\includegraphics[width=0.24\textwidth]{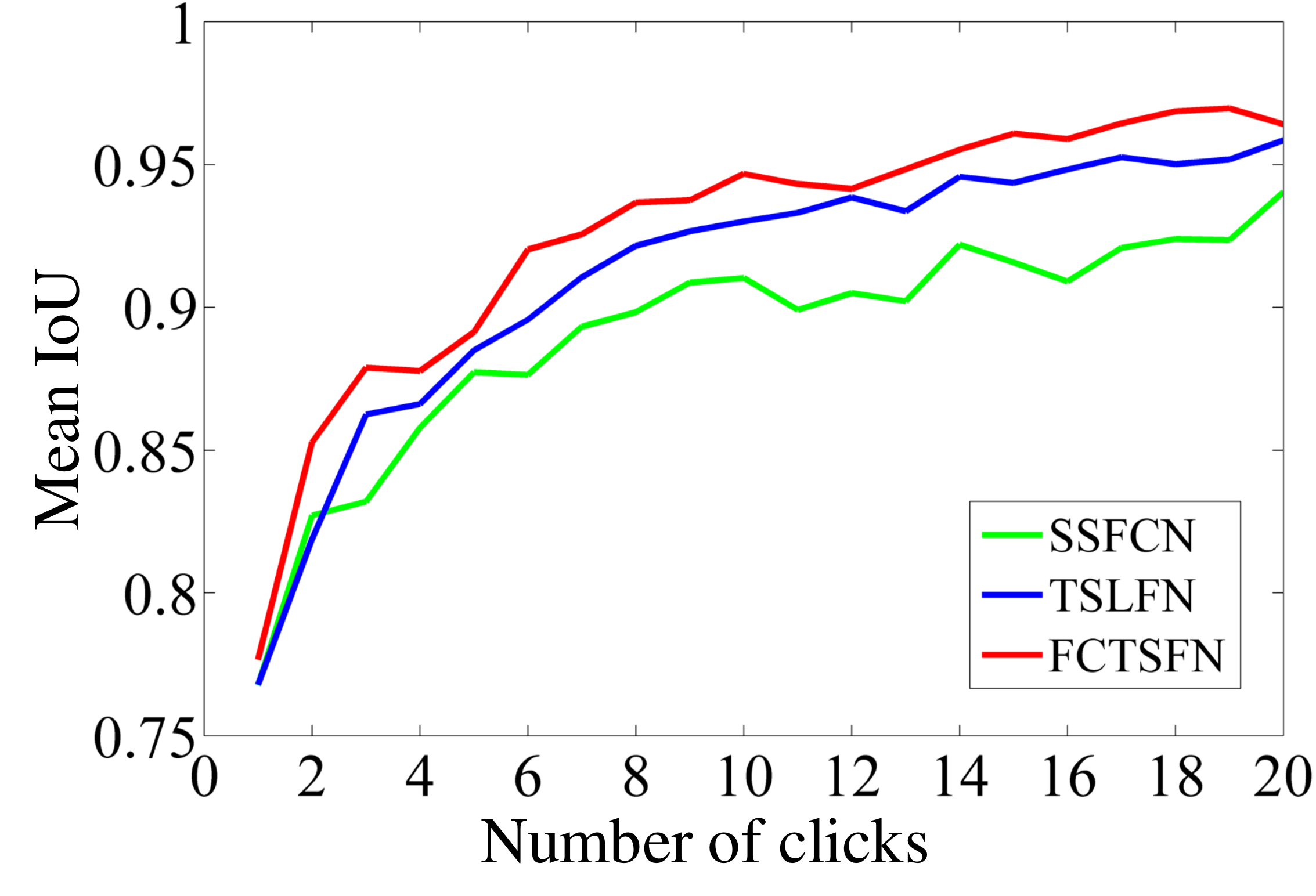}
}
\hspace{-4mm}
\subfigure[Berkeley]
{
\includegraphics[width=0.24\textwidth]{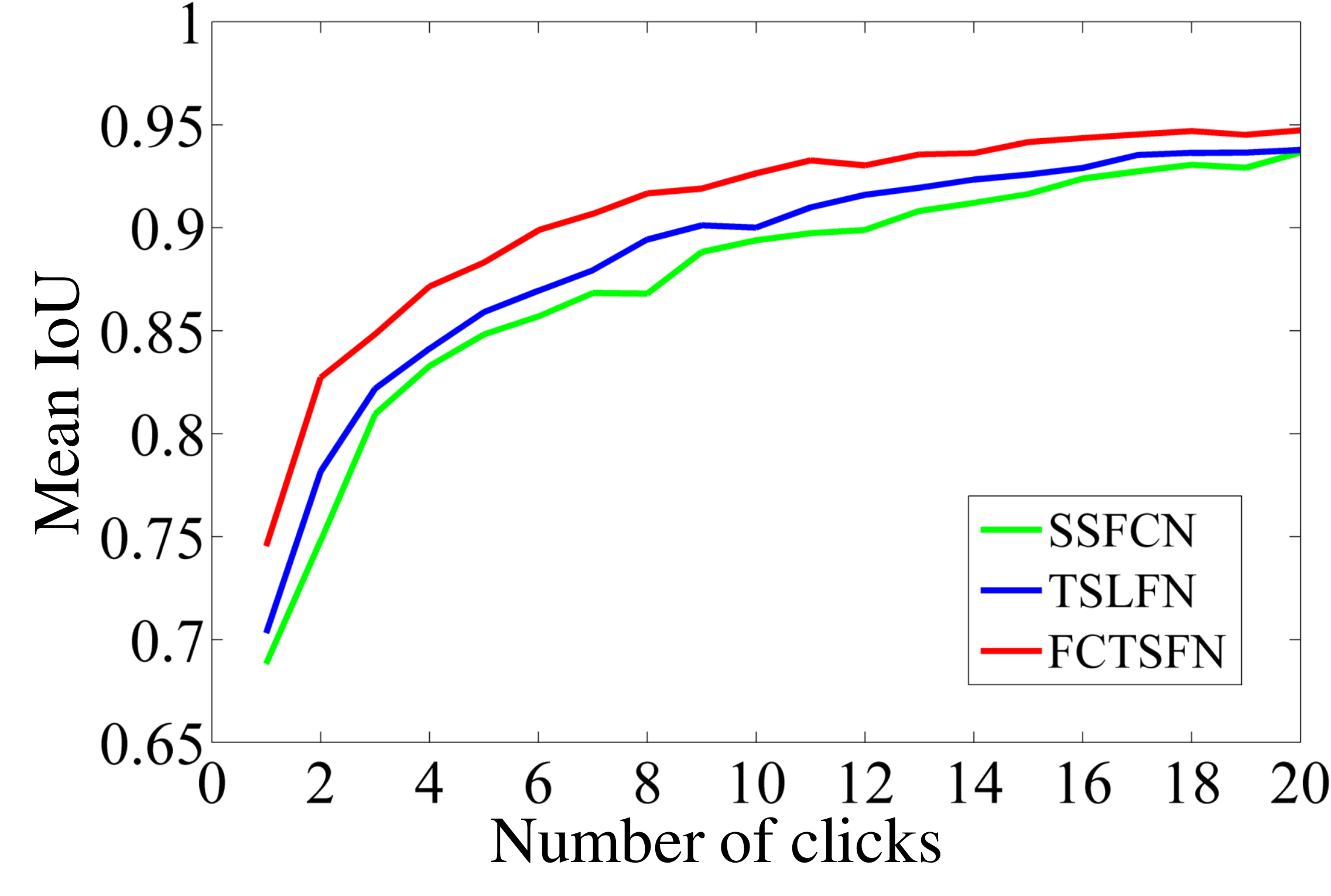}
}
\caption{Mean IoU vs.~number of clicks for the analysis of the TSLFN and MSRN}
\label{fig_experiment_sa}
\end{figure*}

Since the DIs calculated as above are based on the regions around the positive and negative user clicks, they do not represent the separability of the responses between the whole foreground and background regions; hence, they do not represent how good the final segmentation results are. To validate if the higher impact of user clicks benefits the final segmentation performance, we also need to compare the segmentation accuracy between TSLFN and SSFCN. Note that, in the rest of the paper, we do not restrict the types of user clicks (\textit{i.e.}~we follow the free-choice setting above). This is for two reasons. First, it is the most general case to allow users to freely place positive and negative clicks. Second, the free-choice setting actually covers the all-positive and single-positive settings; for example, when the foreground is very small, the free-choice setting is very likely to produce a click sequence the same as the click sequence produced by the single-positive setting. As shown in Fig.~\ref{fig_experiment_sa}, Tab.~\ref{tab_experiment_sa} and Tab.~\ref{tab_experiment_IoU_sa}, TSLFN has a better final segmentation performance compared to SSFCN. This observation suggests that an improved performance is indeed achieved by improving the impact of user interactions on the network output with a two-stream network architecture.

Another interesting observation from Tab.~\ref{tab_experiment_DI_TSLFN} is that the DIs generally drop as the number of user clicks increases. We think there are two possible reasons. First, with a single user click, the network may be more focused around this only click; this leads to very large difference between the responses around the click and the responses in the background, so it leads to a large DI. In contrast, with more user clicks, the network may try to achieve a trade-off between the influence of all clicks; this may lead to decreased responses around each click and hence a lower DI. Second, we find that the main object can be generally segmented with high accuracy with very few clicks (see Tab.~\ref{tab_experiment_sa}). As a result, with 5 or 10 user clicks, there may be many clicks near to object boundary. The responses around these clicks may lower the overall DI, as the responses in the foreground probability maps may be weaker around the object boundary and hence they are less separating between positive and negative clicks.

Moreover, closely examining the results in Tab.~\ref{tab_experiment_DI_TSLFN}, we can find that the above observation on the decreasing DI with respect to the increasing number of user clicks only holds for the free-choice and all-positive settings. For the single-positive setting, the DI is very similar between 5 and 10 clicks. This observation leads to a further interesting possibility on the behaviour of the network: the network treats the positive clicks competitively, while it treats the negative clicks equally. On the one hand, with the all-positive setting, the DIs decrease when the number of clicks increases. This may mean that there exists a competition between the impact of each individual positive click; it leads to a trade-off on the impact of each positive click and this lowers the overall DI with more positive clicks. On the other hand, with the single-positive setting, the DIs are very similar between 5 and 10 clicks. This may mean that adding negative clicks changes little on the impact of each individual negative click. In other words, the network treats each negative click equally.

\begin{table*}[!t]
\begin{center}
\caption{Mean number of clicks to achieve a certain IoU for the analysis of the TSLFN and MSRN (best performance in bold)}
\label{tab_experiment_sa}
\begin{tabular}{cccc}
\hline\noalign{\smallskip}
Dataset & SSFCN & TSLFN & FCTSFN \\
\noalign{\smallskip}
\hline
\noalign{\smallskip}
Pascal VOC 2012 (85\% IoU) & 5.81 & 4.95 & {\bf 4.58} \\
Microsoft Coco (85\% IoU) & 11.42 & 9.97 & {\bf 9.62} \\
Grabcut (90\% IoU) & 5.02 & 4.28 & {\bf 3.76} \\
Berkeley (90\% IoU) & 8.48 & 7.89 & {\bf 6.49} \\
\hline
\end{tabular}
\end{center}
\end{table*}

\begin{table*}[!t]
\begin{center}
\caption{Mean IoU at certain number of clicks for the analysis of the TSLFN and MSRN (in percentage, best performance in bold)}
\label{tab_experiment_IoU_sa}
\begin{tabular}{cccc}
\hline\noalign{\smallskip}
Dataset & SSFCN & TSLFN & FCTSFN \\
\noalign{\smallskip}
\hline
\noalign{\smallskip}
Pascal VOC 2012 (1 click) & 57.0 & 60.0 & {\bf 62.3} \\
Microsoft Coco (1 click) & 38.6 & 42.3 & {\bf 42.5} \\
Grabcut (1 click) & 76.8 & 76.8 & {\bf 77.7} \\
Berkeley (1 click) & 68.8 & 70.3 & {\bf 74.5} \\
\hline
Pascal VOC 2012 (3 clicks) & 73.0 & 76.8 & {\bf 78.0} \\
Microsoft Coco (3 clicks) & 55.1 & 60.7 & {\bf 61.2} \\
Grabcut (3 clicks) & 83.2 & 86.3 & {\bf 87.9} \\
Berkeley (3 clicks) & 81.0 & 82.2 & {\bf 84.8} \\
\hline
Pascal VOC 2012 (10 clicks) & 88.2 & 91.7 & {\bf 92.6} \\
Microsoft Coco (10 clicks) & 72.8 & 80.0 & {\bf 81.5} \\
Grabcut (10 clicks) & 91.0 & 93.0 & {\bf 94.7} \\
Berkeley (10 clicks) & 89.4 & 90.0 & {\bf 92.6} \\
\hline
\end{tabular}
\end{center}
\end{table*}

The above experiments validate our idea that the two-stream network allows the user clicks to have a higher impact on the prediction result, and it leads to improved performance compared to the single-stream network. However, this leads to another question: do we need deep features of user interactions to apply this impact? In other words, is the interaction stream necessary for the TSLFN? To justify the effect of the interaction stream that produces deep features for user interactions, we compare the performance between two networks: (1) TSLFN; (2) TSLFN with the interactive stream removed (referred to as TSLFN-). Specifically, in TSLFN-, the interaction maps are resized and concatenated with the features at the end of the image stream; the concatenated features are then used as the input of the fusion net to predict the foreground (note that this is different from SSFCN where the interaction maps are concatenated with the original input image at the beginning of the whole network). Fig.~\ref{fig_experiment_ablation_analysis} compares the performance between TSLFN- and TSLFN (note that the results in this figure are based on the stride-32 networks; the performance on the final stride-8 networks are likely to be similar, as the stride-8 networks are based on the stride-32 networks). It can be seen that the performance drops for TSLFN-. This result shows that the deep features of user interactions from the interaction stream is also important for the TSLFN to achieve a good performance. This may be because deep features provide a richer and more meaningful representation of user clicks, and it can more accurately guide the segmentation process when fused with image features.

Finally, we report some experimental results related to the design of the proposed TSLFN. As discussed at the end of section~\ref{sec:proposed_method_TSLFN}, we could use different depth in the image/interaction streams and the fusion net to construct TSLFN, given the VGG16 base network. Specifically, the VGG16 base network has 5 Conv-ReLU-Pool (CRP) blocks. Our TSLFN structure in Fig.~\ref{fig_TSLFN} uses the first 4 CRP blocks to form the image/interaction streams, and it uses the rest part of VGG16 as the fusion net. One can create variations of this architecture by using a different number of CRB blocks to form the image/interaction streams. This leads to variations of the proposed TSLFN with different depth in the image/interaction streams and the fusion net. We use TSLFN\_$i$ to denote the variation of the proposed TSLFN with the first $i$ CRP blocks in the base network used as the image/interaction streams. The proposed TSLFN shown in Fig.~\ref{fig_TSLFN} is essentially equivalent to TSLFN\_4. It has four variations: TSLFN\_1, TSLFN\_2, TSLFN\_3, TSLFN\_5. Among these variations, TSLFN\_1 has the shallowest image/interaction streams and the deepest fusion net, while TSLFN\_5 has the deepest image/interaction streams and the shallowest fusion net.

\begin{figure*}[!t]
\centering
\subfigure[Pascal VOC 2012]
{
\includegraphics[width=0.24\textwidth]{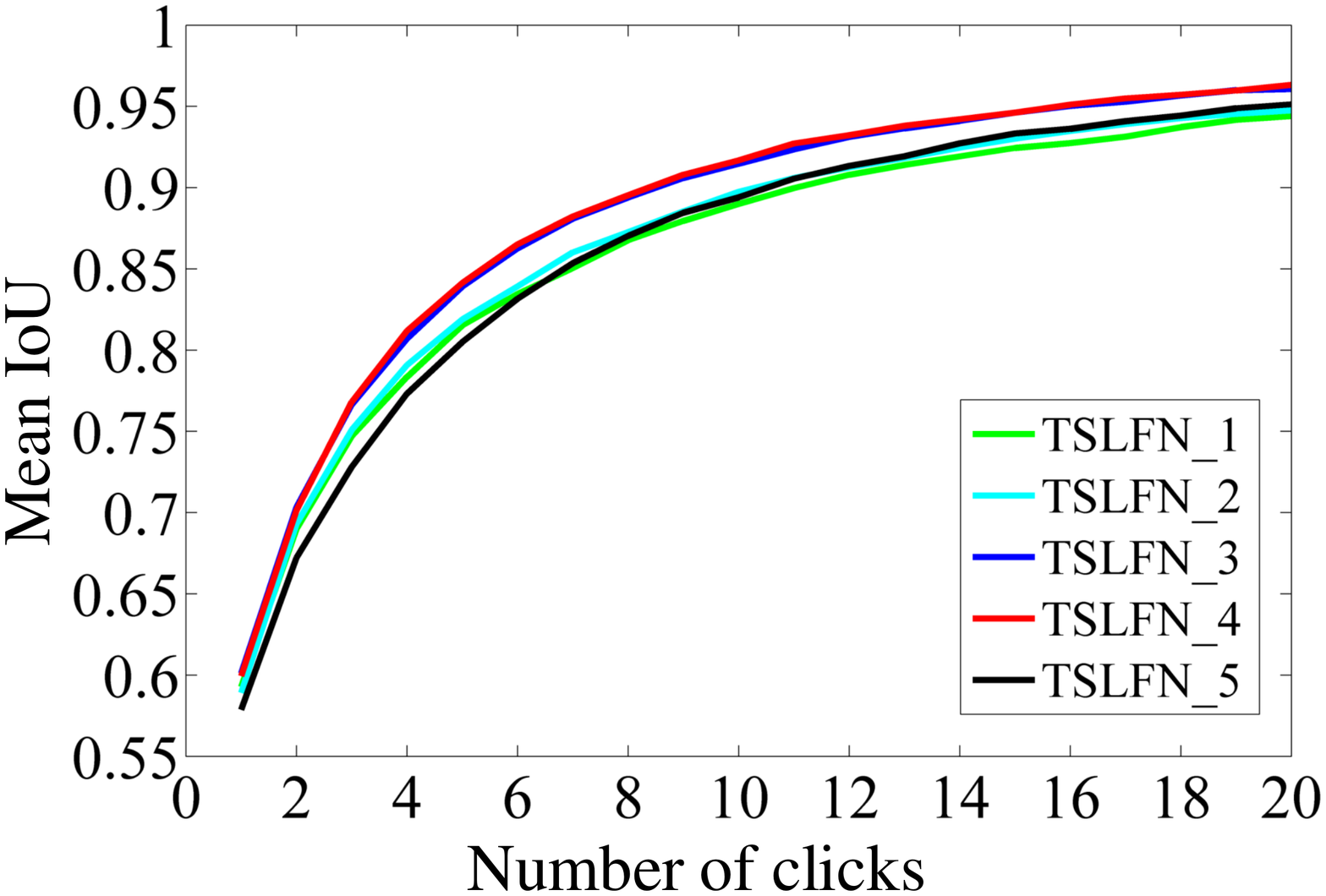}
}
\hspace{-4mm}
\subfigure[Microsoft Coco]
{
\includegraphics[width=0.24\textwidth]{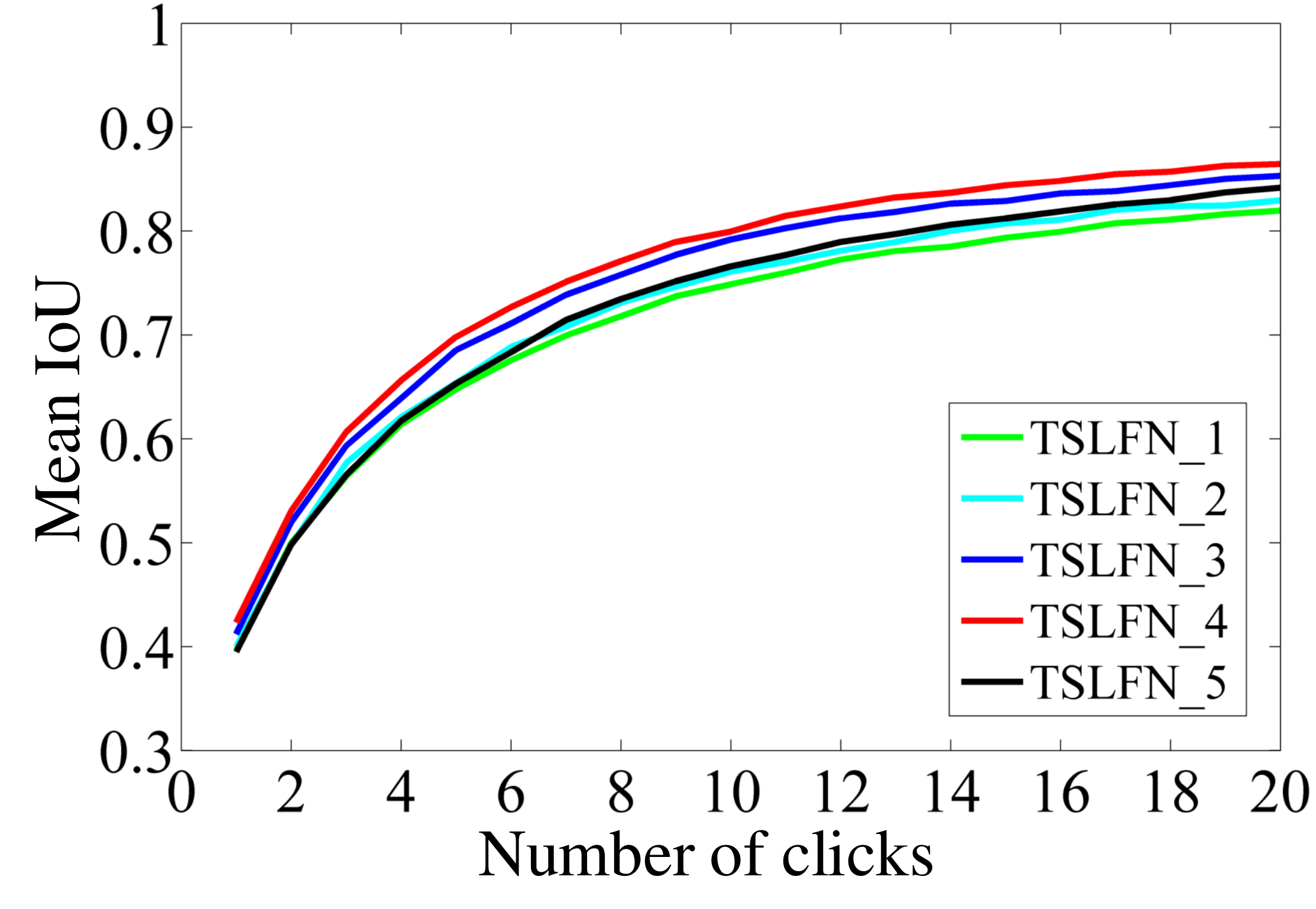}
}
\hspace{-4mm}
\subfigure[Grabcut]
{
\includegraphics[width=0.24\textwidth]{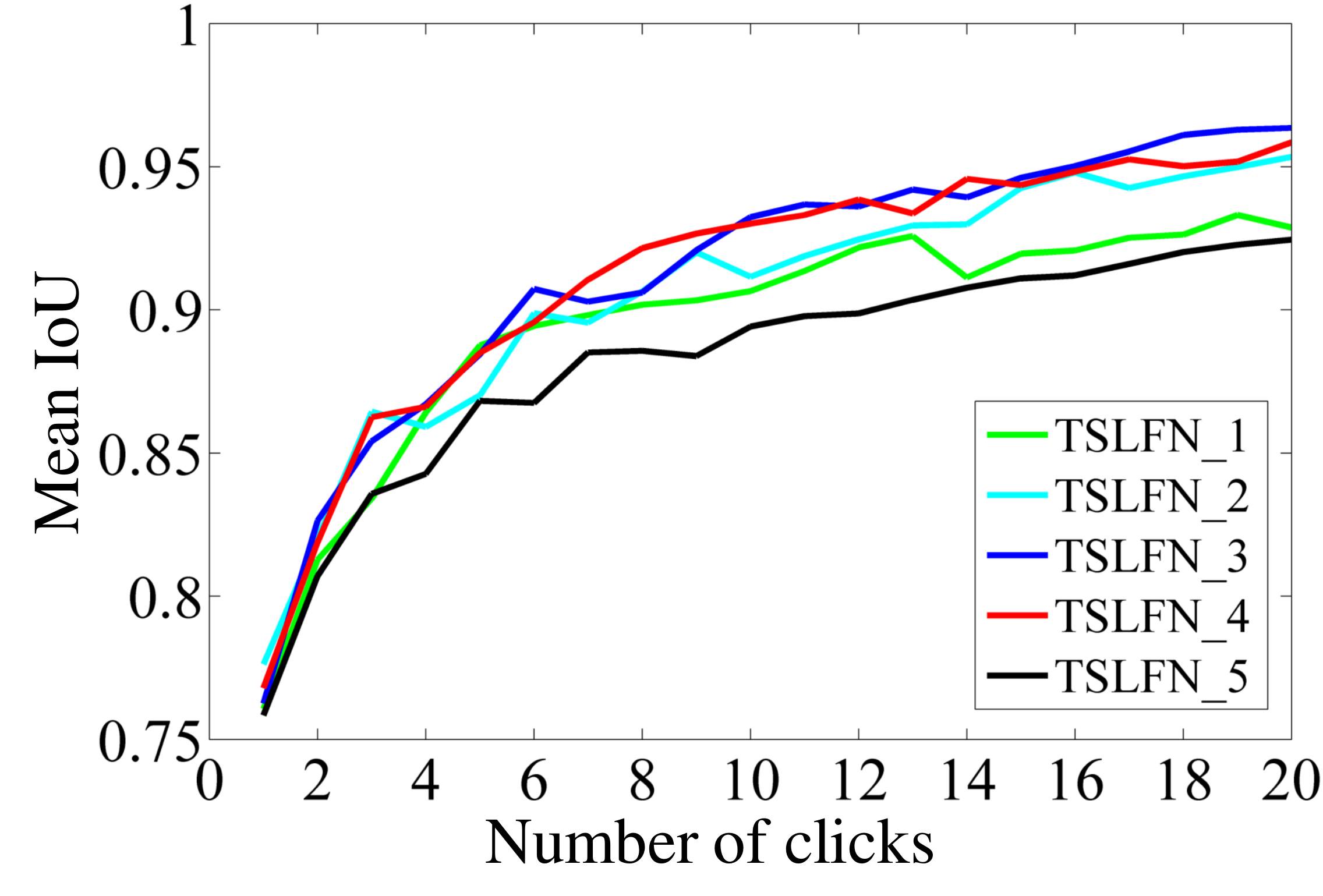}
}
\hspace{-4mm}
\subfigure[Berkeley]
{
\includegraphics[width=0.24\textwidth]{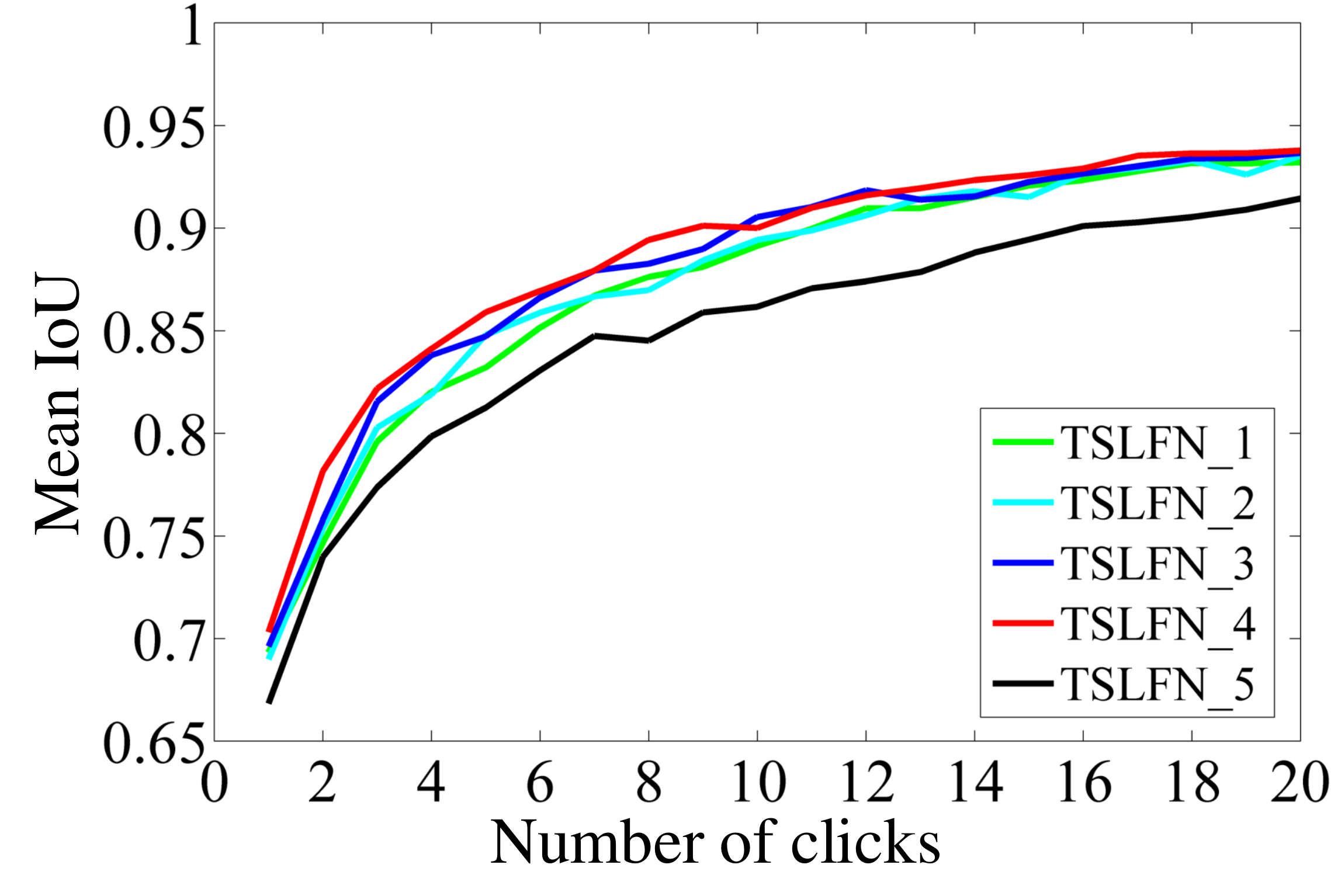}
}
\caption{Mean IoU vs.~number of clicks for the analysis of the design of the proposed TSLFN}
\label{fig_experiment_design_choice_TSLFN}
\end{figure*}

\begin{table*}[!t]
\begin{center}
\caption{Mean number of clicks to achieve a certain IoU for the analysis of the design of the proposed TSLFN (best performance in bold)}
\label{tab_experiment_design_choice_TSLFN}
\begin{tabular}{cccccc}
\hline\noalign{\smallskip}
Dataset & TSLFN\_1 & TSLFN\_2 & TSLFN\_3 & TSLFN\_4 & TSLFN\_5 \\
\noalign{\smallskip}
\hline
\noalign{\smallskip}
Pascal VOC 2012 (85\% IoU) & 5.63 & 5.43 & {\bf 4.95} & {\bf 4.95} & 5.73 \\
Microsoft Coco (85\% IoU) & 11.06 & 10.83 & 10.18 & {\bf 9.97} & 10.99 \\
Grabcut (90\% IoU) & 4.60 & 4.66 & 4.44 & {\bf 4.28} & 5.14 \\
Berkeley (90\% IoU) & 8.21 & 8.57 & 8.15 & {\bf 7.89} & 9.27 \\
\hline
\end{tabular}
\end{center}
\end{table*}

\begin{table*}[!t]
\begin{center}
\caption{Mean IoU at certain number of clicks for the analysis of the design of the proposed TSLFN (in percentage, best performance in bold)}
\label{tab_experiment_design_choice_TSLFN_IoU}
\begin{tabular}{cccccc}
\hline\noalign{\smallskip}
Dataset & TSLFN\_1 & TSLFN\_2 & TSLFN\_3 & TSLFN\_4 & TSLFN\_5 \\
\noalign{\smallskip}
\hline
\noalign{\smallskip}
Pascal VOC 2012 (1 click) & 59.3 & 58.9 & {\bf 60.1} & 60.0 & 57.9 \\
Microsoft Coco (1 click) & 39.7 & 40.0 & 41.2 & {\bf 42.3} & 39.5 \\
Grabcut (1 click) & 76.1 & {\bf 77.6} & 76.3 & 76.8 & 75.9 \\
Berkeley (1 click) & 69.3 & 69.0 & 69.6 & {\bf 70.3} & 66.8 \\
\hline
Pascal VOC 2012 (3 clicks) & 74.7 & 75.1 & 76.7 & {\bf 76.8} & 72.8 \\
Microsoft Coco (3 clicks) & 56.4 & 57.7 & 59.3 & {\bf 60.7} & 56.5 \\
Grabcut (3 clicks) & 83.4 & {\bf 86.4} & 85.4 & 86.3 & 83.6 \\
Berkeley (3 clicks) & 79.6 & 80.3 & 81.6 & {\bf 82.2} & 77.4 \\
\hline
Pascal VOC 2012 (10 clicks) & 89.0 & 89.8 & 91.5 & {\bf 91.7} & 89.4 \\
Microsoft Coco (10 clicks) & 74.9 & 76.1 & 79.2 & {\bf 80.0} & 76.6 \\
Grabcut (10 clicks) & 90.7 & 91.2 & {\bf 93.2} & 93.0 & 89.4 \\
Berkeley (10 clicks) & 89.1 & 89.4 & {\bf 90.6} & 90.0 & 86.2 \\
\hline
\end{tabular}
\end{center}
\end{table*}

Fig.~\ref{fig_experiment_design_choice_TSLFN}, Tab.~\ref{tab_experiment_design_choice_TSLFN} and Tab.~\ref{tab_experiment_design_choice_TSLFN_IoU} show the performance of all the above variations of the proposed TSLFN. It can be seen that the proposed TSLFN (TSLFN\_4 in Fig.~\ref{fig_experiment_design_choice_TSLFN}, Tab.~\ref{tab_experiment_design_choice_TSLFN} and Tab.~\ref{tab_experiment_design_choice_TSLFN_IoU}) generally has the highest performance among its variations. A possible reason is as the one we discussed at the end of section~\ref{sec:proposed_method_TSLFN}: there exists a trade-off between the impact of user interactions and the prediction capacity with different depths in image/interaction streams and fusion net; the proposed TSLFN structure as shown in Fig.~\ref{fig_TSLFN} achieves the best trade-off between the two factors compared to its other variations, given our base network.

In this subsection we analysed the proposed TSLFN. The results confirm that: (1) compared to the single-stream network, the two-stream structure of the TSLFN allows the information from user clicks to have a higher impact on the network output, and it leads to better performance; (2) extracting deep features from user interactions is also important for the TSLFN to achieve a better performance. We also validated the design choice of the proposed TSLFN, showing that it generally achieves the best performance among its variations with the given base network.

\subsection{Analysis of MSRN}

To analyse the effect of the MSRN, we compare the performance between two networks: TSLFN and FCTSFN (\textit{i.e.}~TSLFN+MSRN). By comparing the performance reported in Fig.~\ref{fig_experiment_sa}, Tab.~\ref{tab_experiment_sa} and Tab.~\ref{tab_experiment_IoU_sa}, we can see that FCTSFN has a consistently better performance than TSLFN. These observations validate the effectiveness of the MSRN to utilize multi-scale features to refine the segmentation result. In our opinion, two possible reasons lead to the improved performance. First, MSRN makes prediction at full resolution, hence it is more accurate at object boundaries. Second, MSRN utilizes features from the beginning to the end of the network. Therefore, it fuses information from low-level features such as colors/boundaries to high-level features with object-level understanding; this allows the network to build a more comprehensive understanding on the foreground and background, and it leads to more accurate segmentation results.

\subsection{Comparison with existing algorithms}
\label{sec:experiments_comparison}

In this subsection, we compare the proposed network to state-of-the-art algorithms. We divide our comparisons into two categories: restricted comparison and unrestricted comparison. In the restriction comparison, we conduct experiments strictly under our experimental setting in section~\ref{sec:experiments_setting}; we either run the source codes of the comparison methods or implement the comparison methods by ourselves. In the unrestricted comparison, we directly compared with the performance measure cited from published papers. Note that, in the unrestricted comparison, the results are not fully comparable due to the differences in the experimental setting in different papers. However, it shows the performance of the proposed network among state-of-the-art methods with open choices for experimental settings.

{\bf Restricted comparison.} For the restricted comparison, we compare to the following methods: graph cut (GC)~\cite{GraphCut_Boykov_2001}, geodesic matting (GM)~\cite{GeodesicMatting_Bai_2007}, random walk (RW)~\cite{RandomWalk_Grady_2006}, Euclidean star convexity (ESC)~\cite{GeodesicStar_Gulshan_2010}, geodesic star convexity (GSC)~\cite{GeodesicStar_Gulshan_2010}, single stream FCN (SSFCN)~\cite{FCN_InteractiveSeg_Xu_2016}.

\begin{figure*}[!t]
\centering
\subfigure[Pascal VOC 2012]
{
\includegraphics[width=0.24\textwidth]{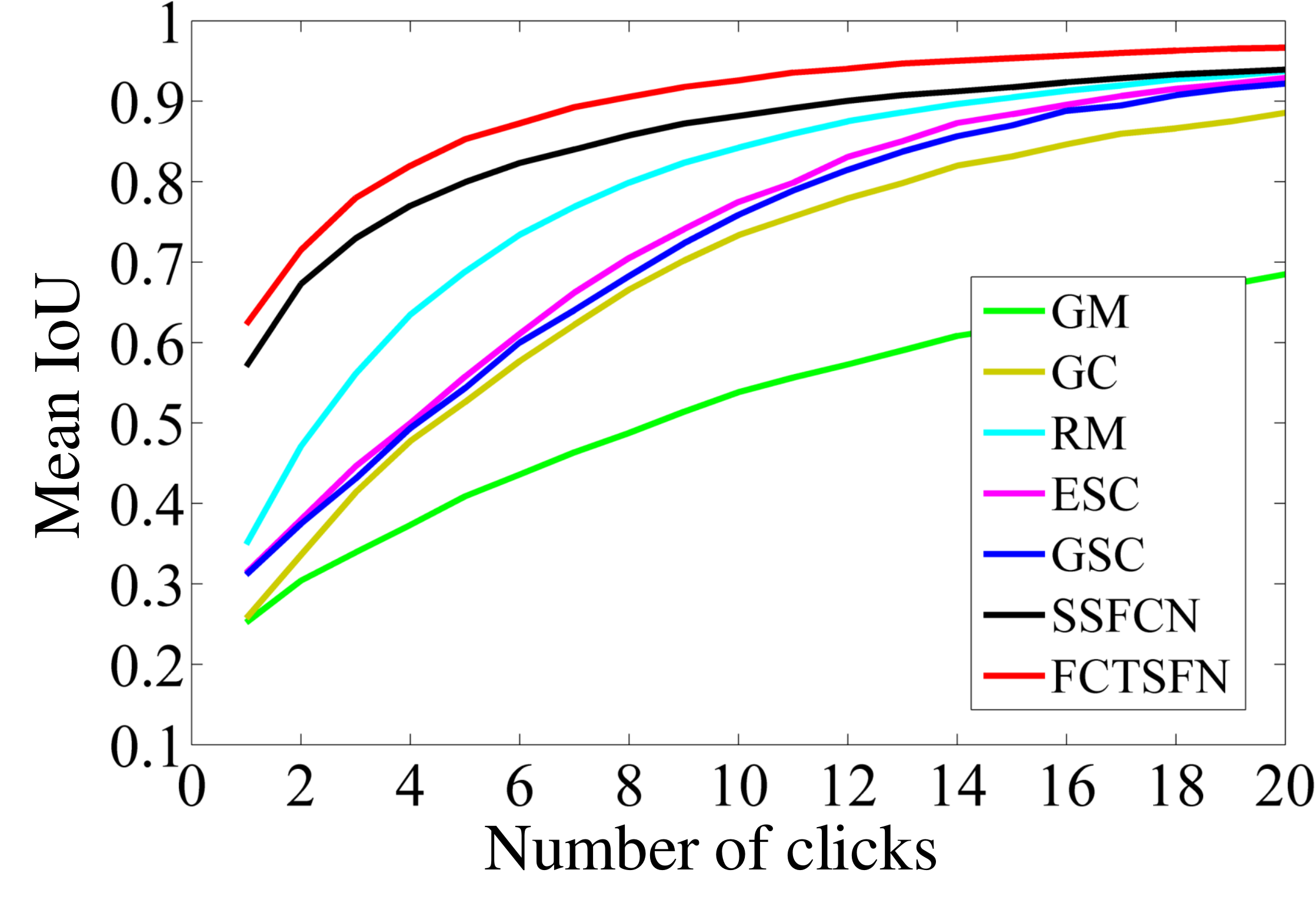}
}
\hspace{-4mm}
\subfigure[Microsoft Coco]
{
\includegraphics[width=0.24\textwidth]{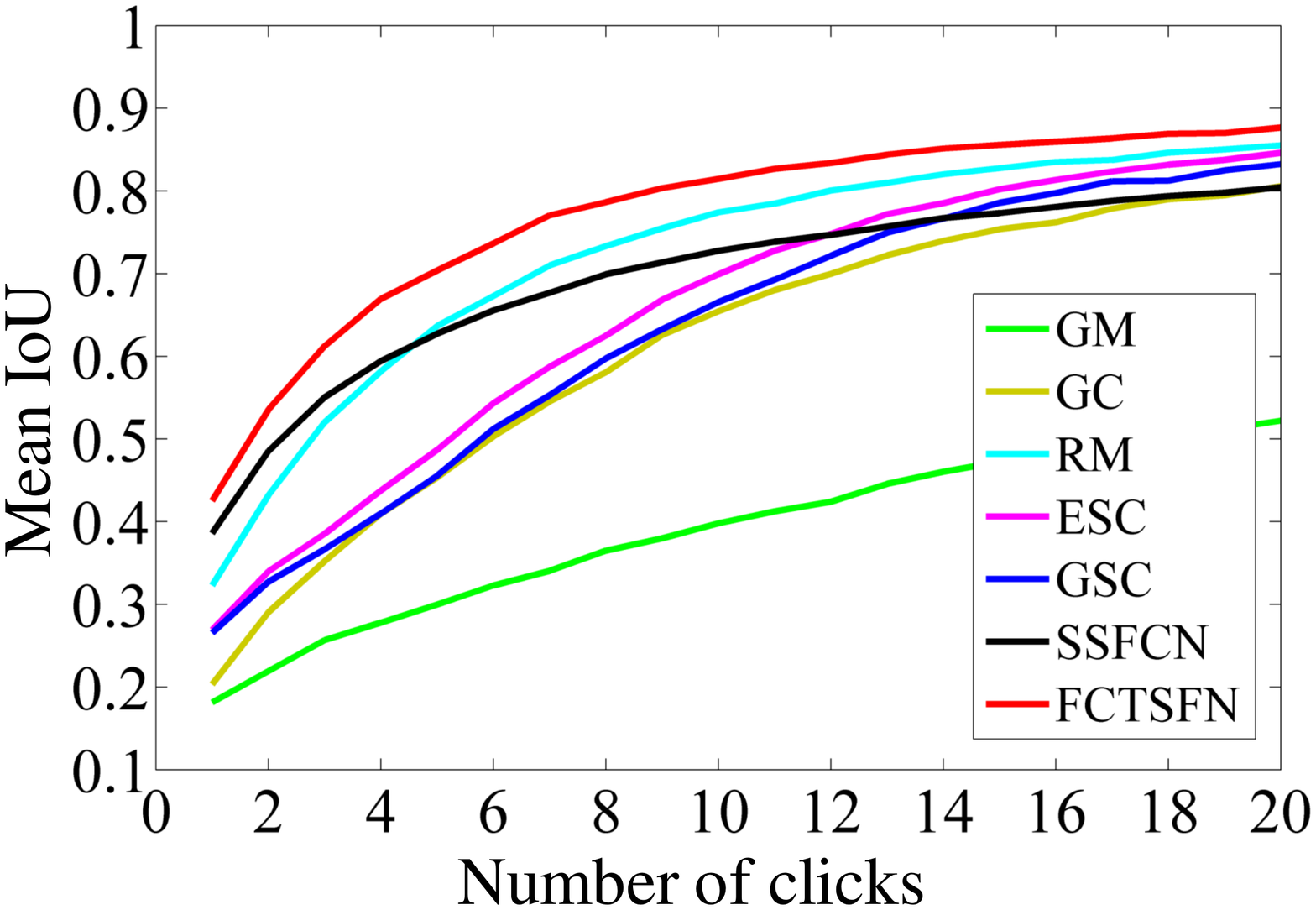}
}
\hspace{-4mm}
\subfigure[Grabcut]
{
\includegraphics[width=0.24\textwidth]{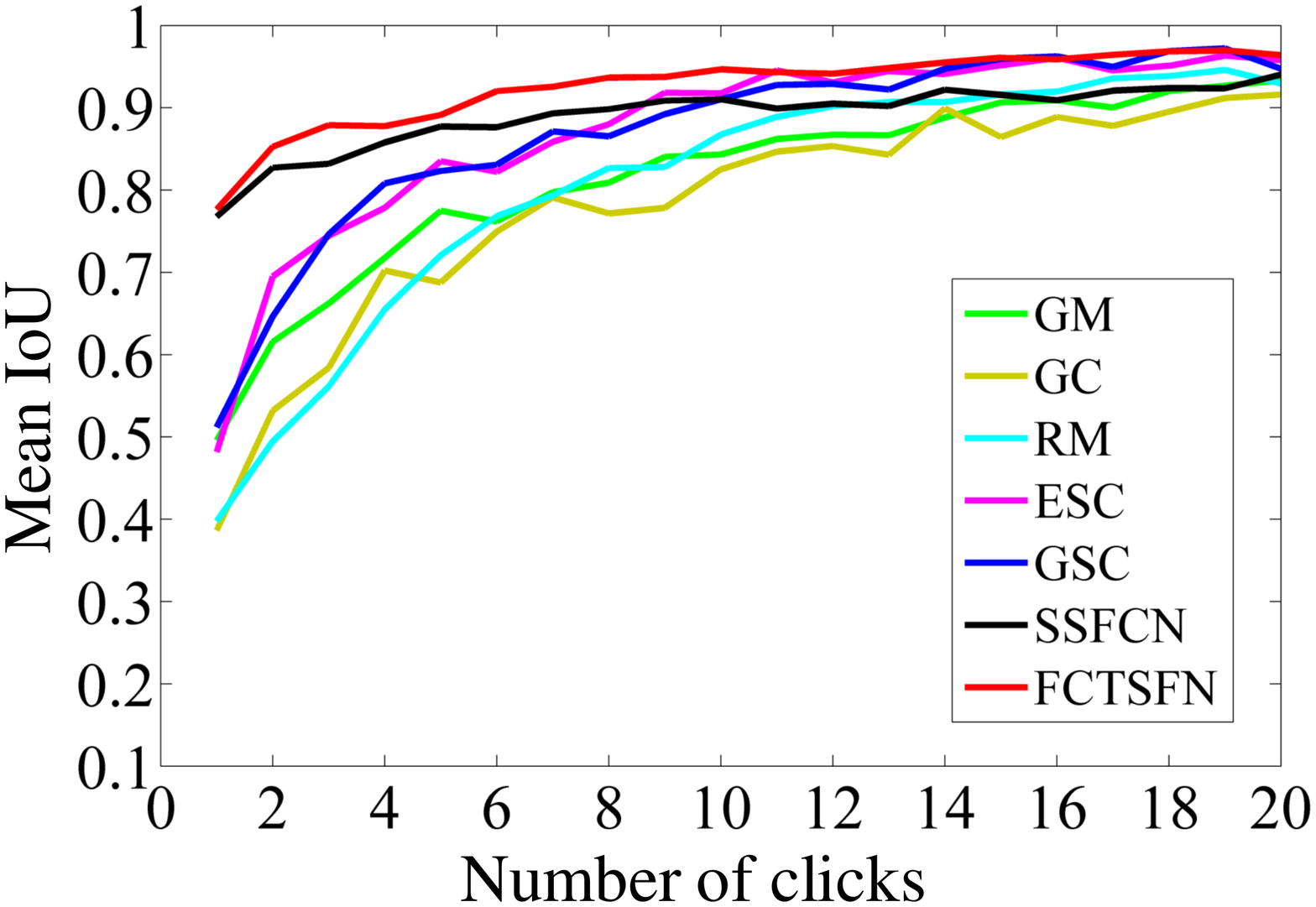}
}
\hspace{-4mm}
\subfigure[Berkeley]
{
\includegraphics[width=0.24\textwidth]{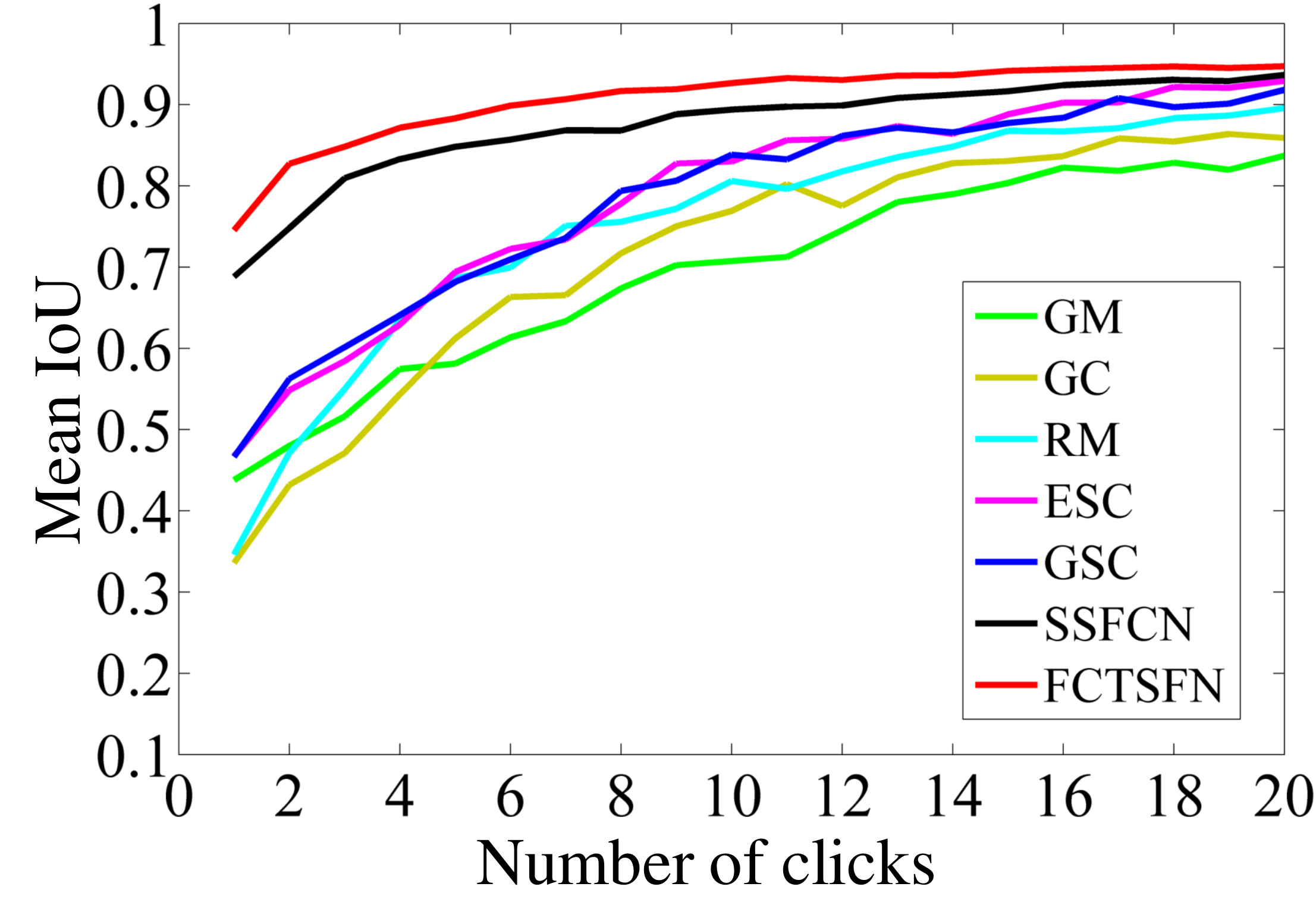}
}
\caption{Mean IoU vs.~number of clicks for restricted comparisons}
\label{fig_experiment_cmp}
\end{figure*}

\begin{figure*}[t]
\centering
\subfigure[]
{
\includegraphics[width=0.107\textwidth]{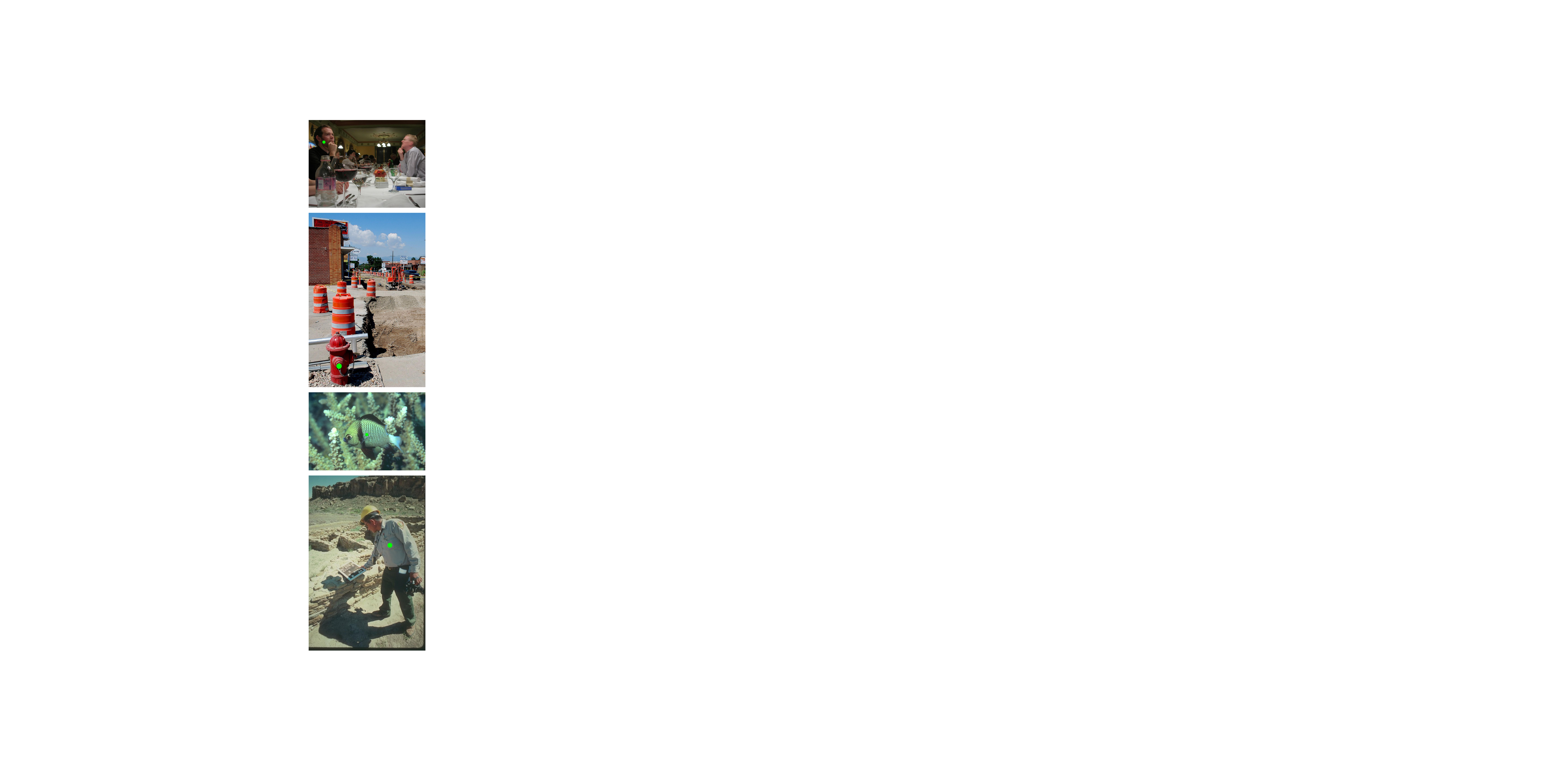}
}
\hspace{-1.2em}
\subfigure[]
{
\includegraphics[width=0.107\textwidth]{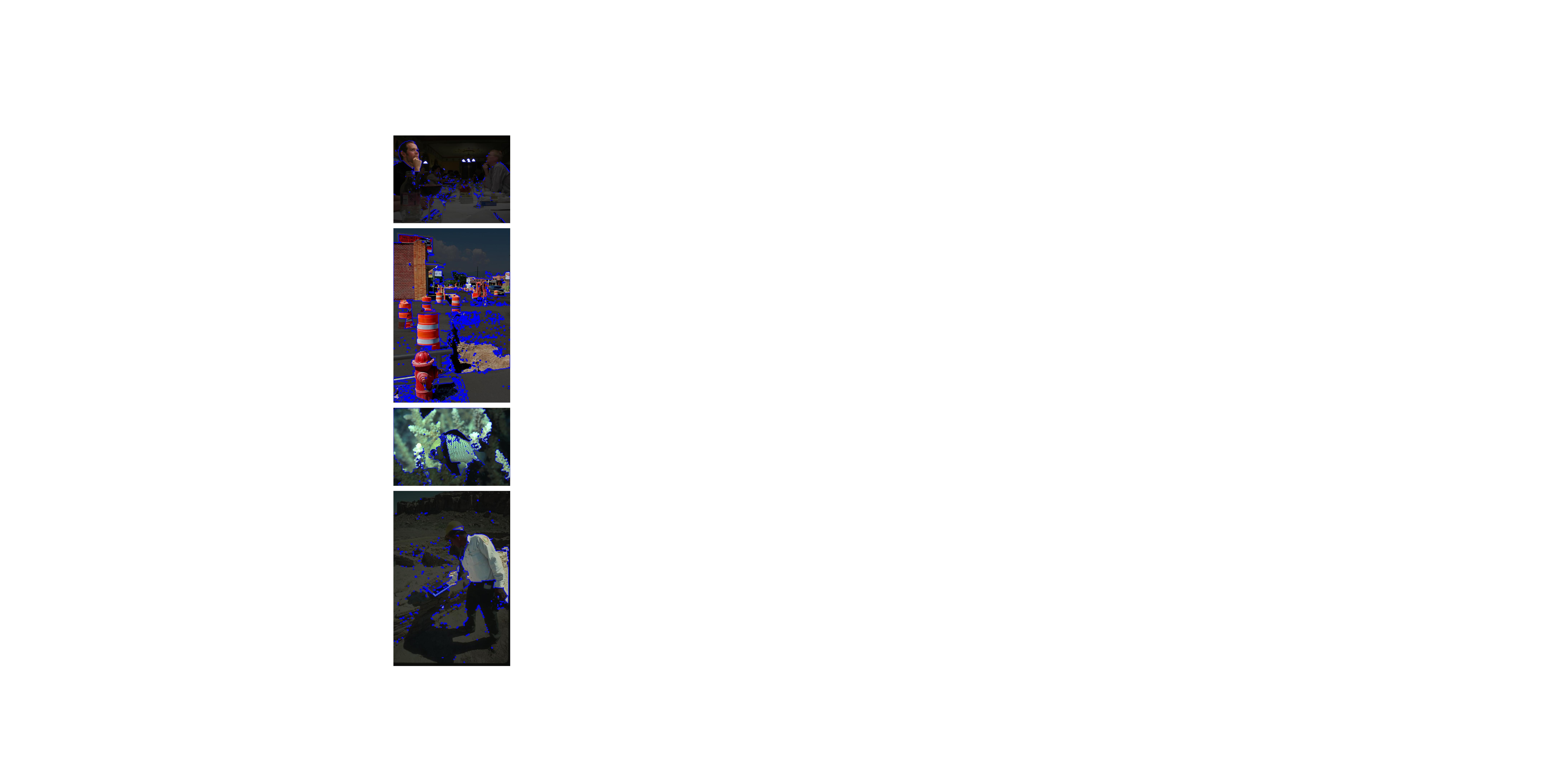}
}
\hspace{-1.2em}
\subfigure[]
{
\includegraphics[width=0.107\textwidth]{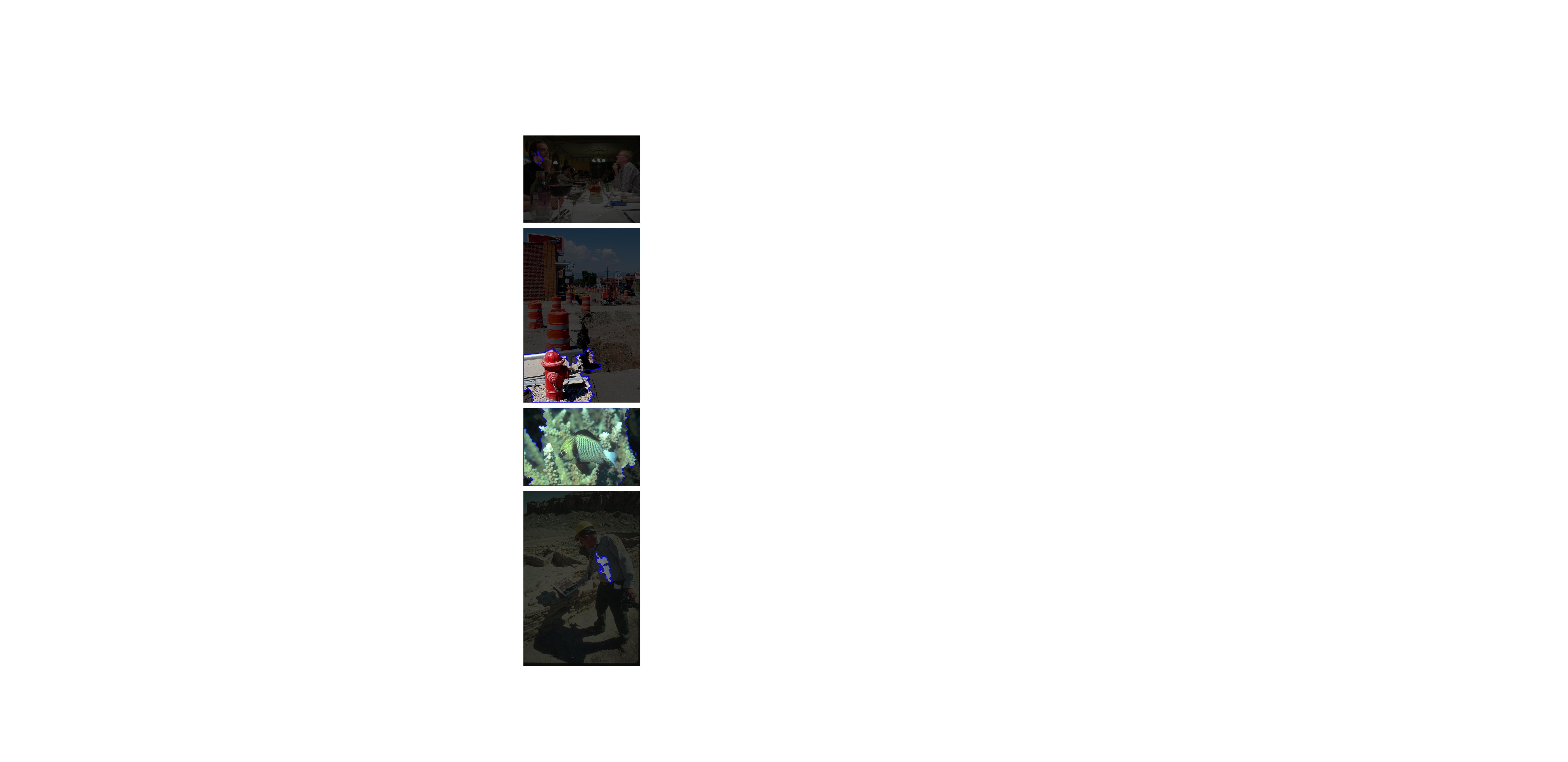}
}
\hspace{-1.2em}
\subfigure[]
{
\includegraphics[width=0.107\textwidth]{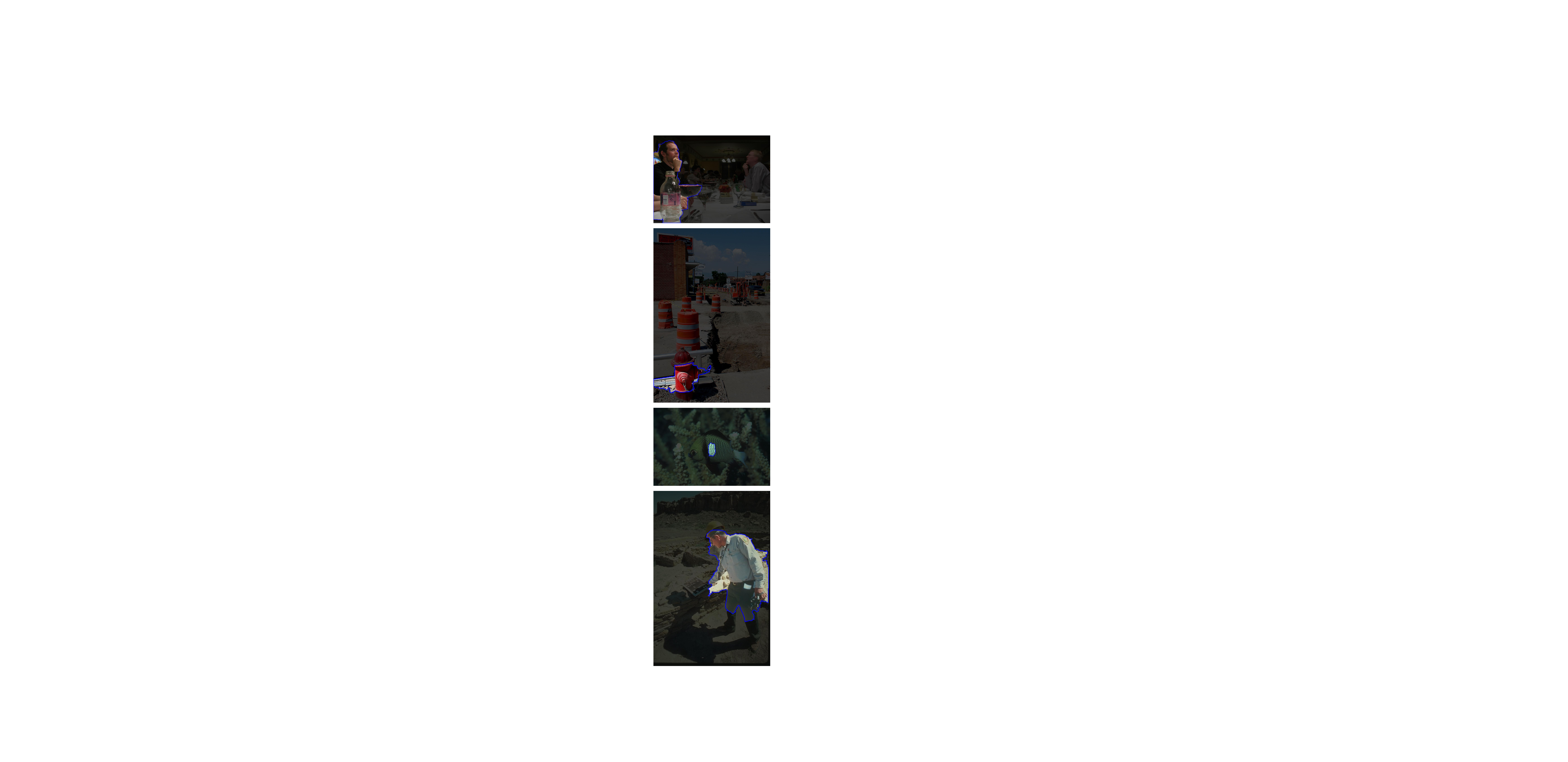}
}
\hspace{-1.2em}
\subfigure[]
{
\includegraphics[width=0.107\textwidth]{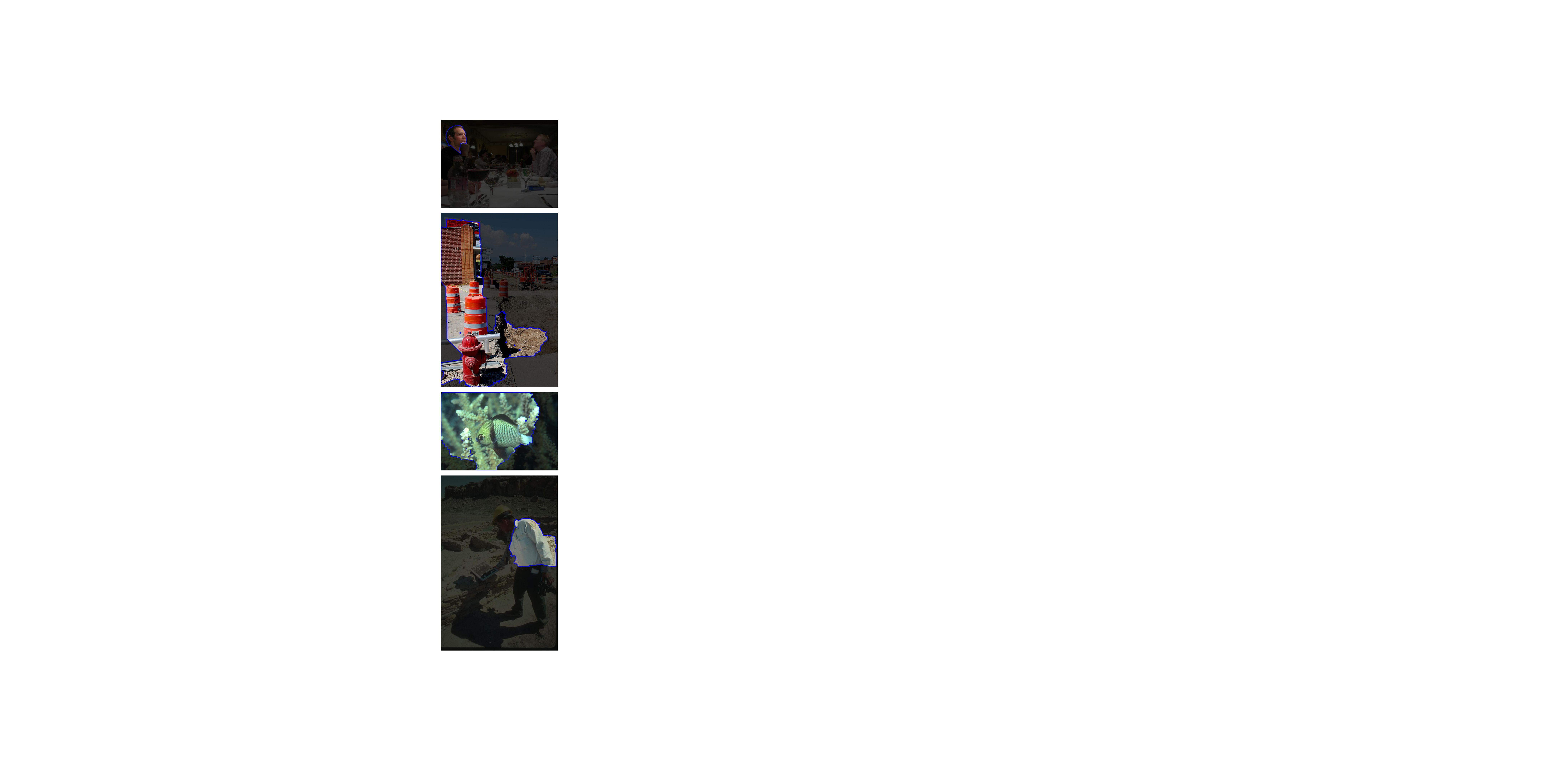}
}
\hspace{-1.2em}
\subfigure[]
{
\includegraphics[width=0.107\textwidth]{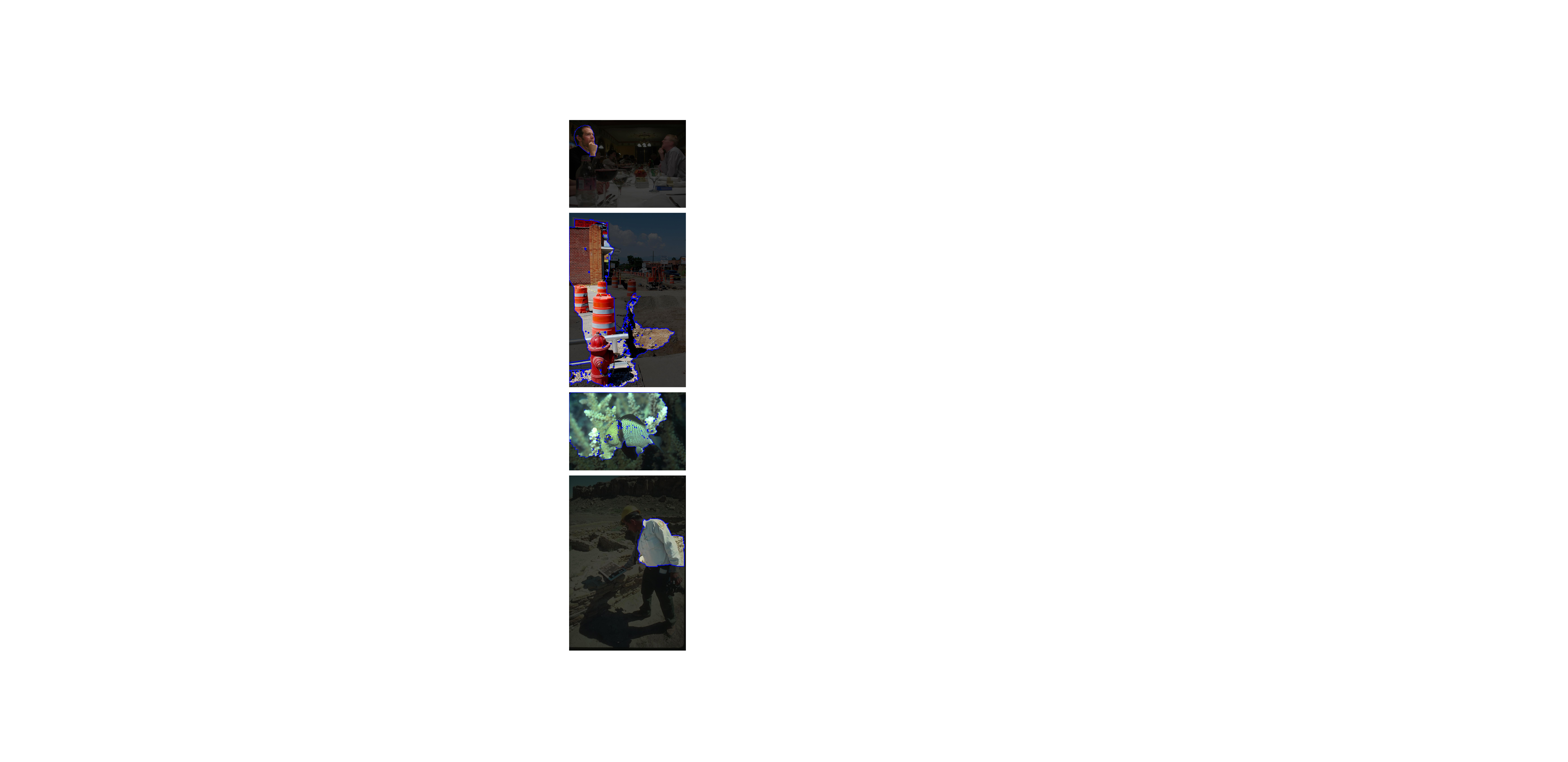}
}
\hspace{-1.2em}
\subfigure[]
{
\includegraphics[width=0.107\textwidth]{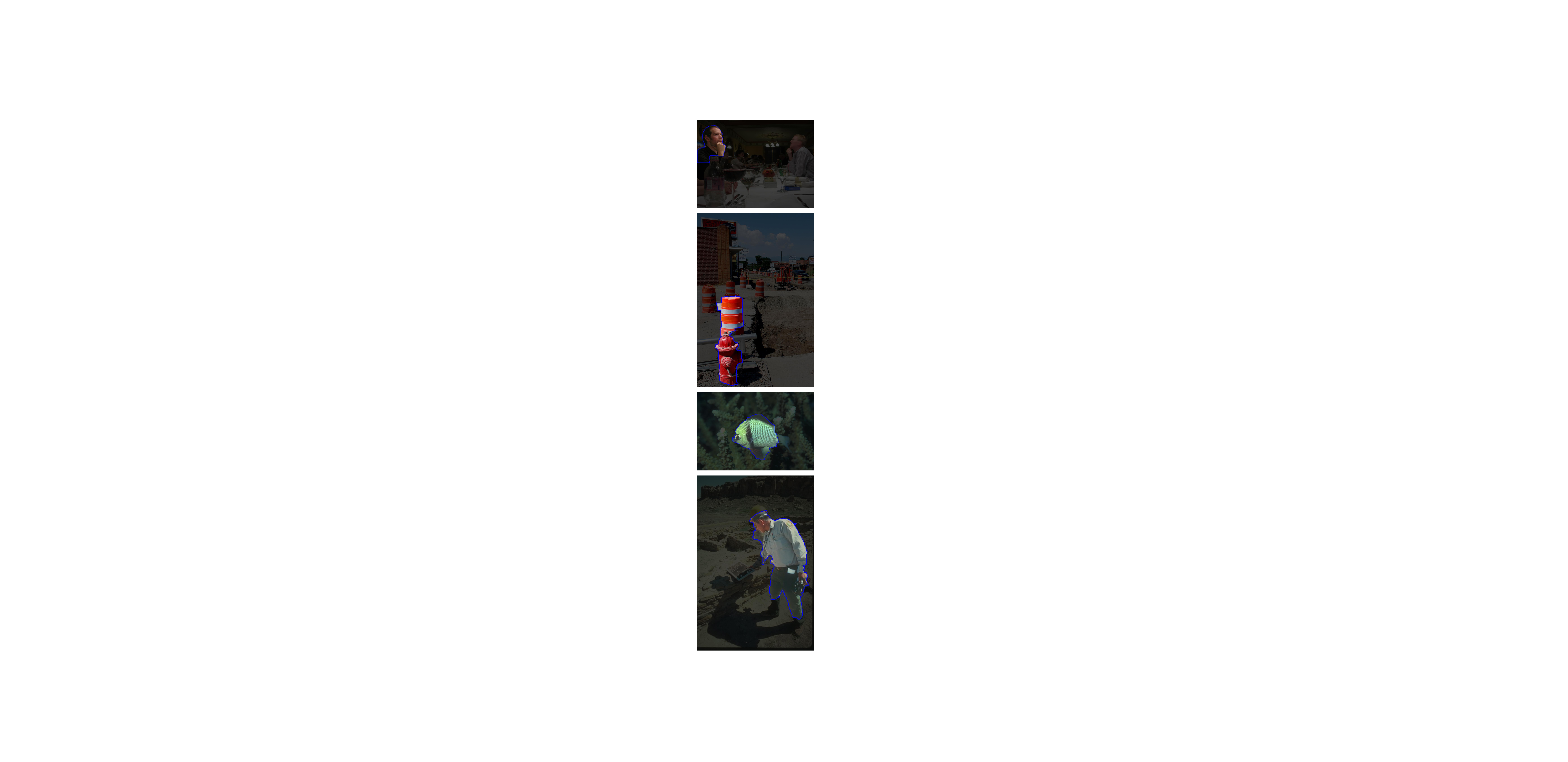}
}
\hspace{-1.2em}
\subfigure[]
{
\includegraphics[width=0.107\textwidth]{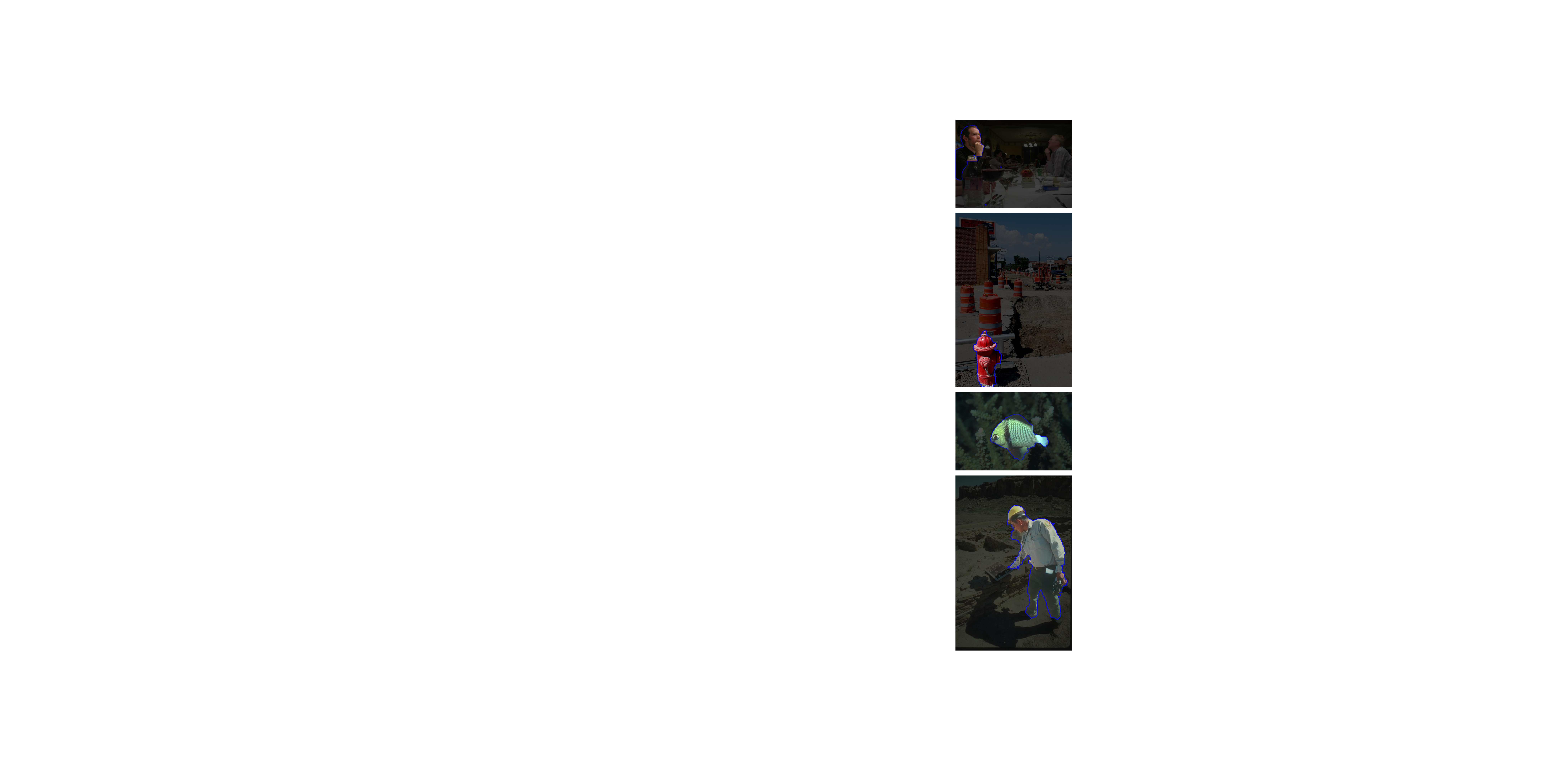}
}
\hspace{-1.2em}
\subfigure[]
{
\includegraphics[width=0.107\textwidth]{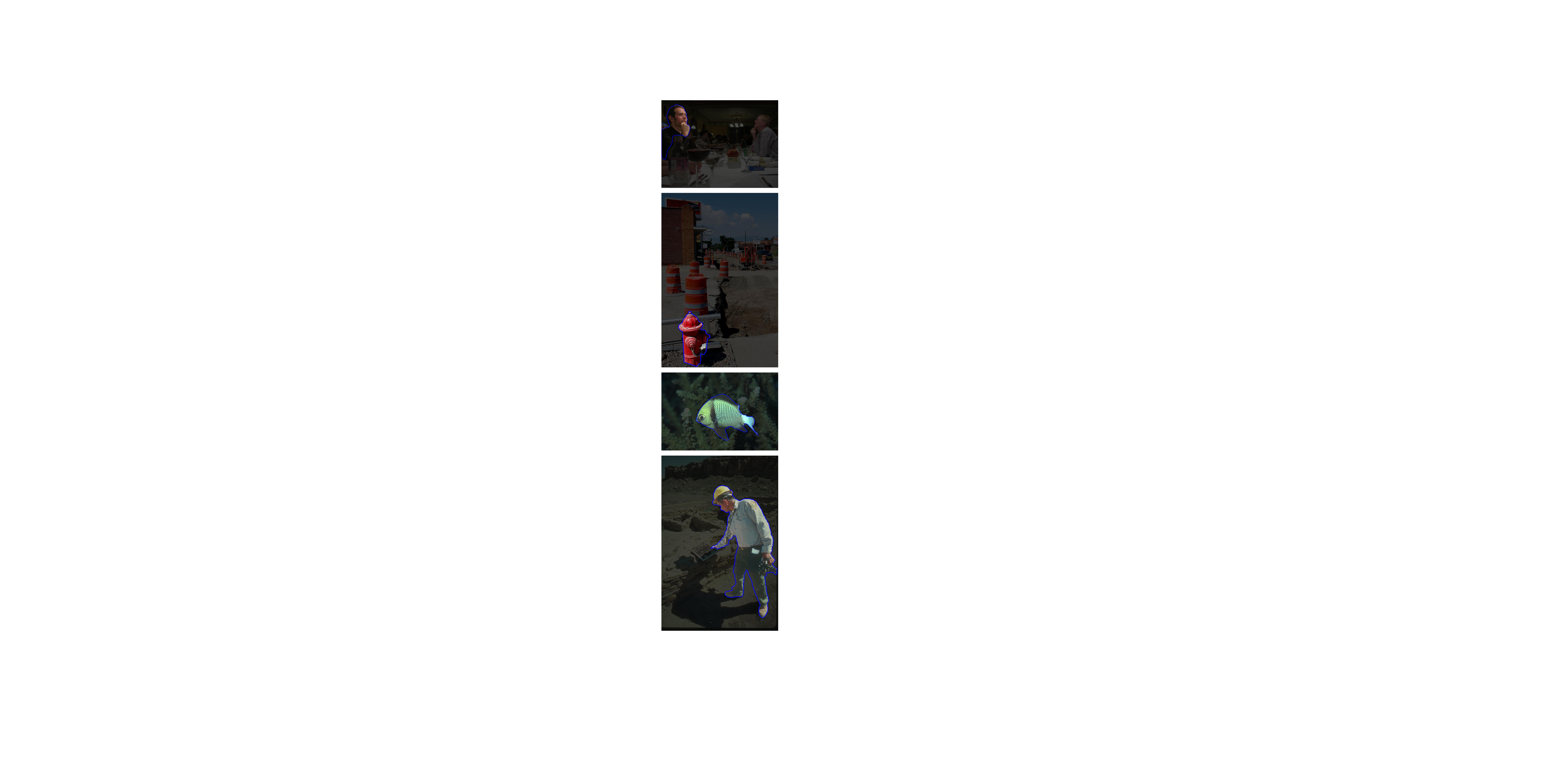}
}
\caption{Examples of segmentation results of different methods given the same user interaction. (a) Original image with user clicks; (b) GC; (c) GM; (d) RW; (e) ESC; (f) GSC; (g) SSFCN; (h) FCTSFN; (i) Ground truth}
\label{fig_example_rst}
\end{figure*}

\begin{figure*}[t]
\centering
\includegraphics[width=0.7\textwidth]{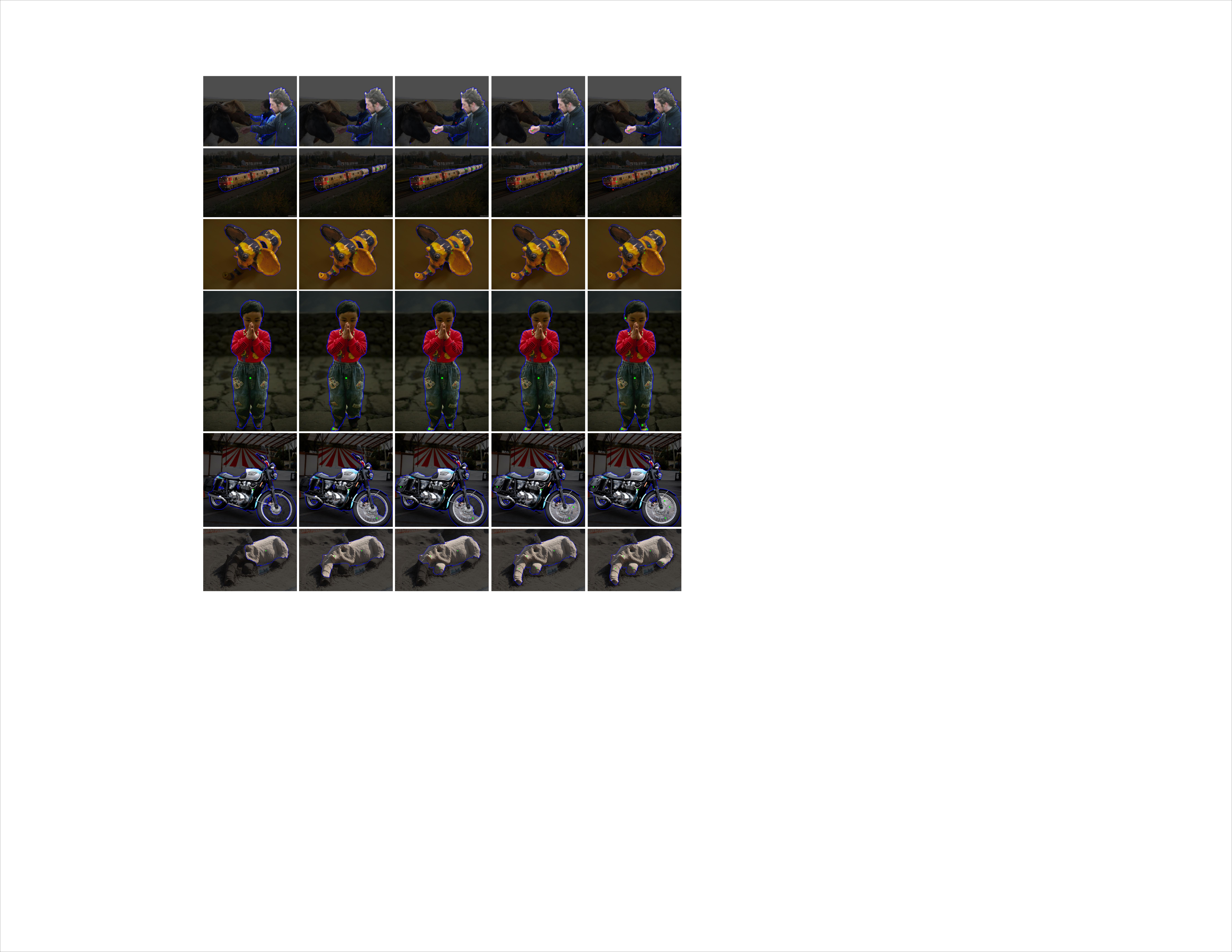}
\caption{Examples of segmentation results of the proposed method on different objects in the test data with automatically generated click sequences; from left to right in each row: the number of clicks increases from 1 to 5.}
\label{fig_example_rst_auto_clicks}
\end{figure*}

Fig.~\ref{fig_experiment_cmp} shows the mean IoU vs.~number of clicks for all comparison methods for all the four datasets on the test data. Tab.~\ref{tab_experiment_IoU_cmp} shows the mean IoU at some certain number of clicks (1, 3, 10). It can be seen that the proposed FCTSFN achieves improved performance compared to the other methods. Specifically, on the Pascal VOC 2012, Microsoft Coco and Berkeley datasets, FCTSFN performs better compared to the other methods. On the Grabcut dataset, FCTSFN achieves better performance when the number of clicks is lower than 10; when the number of clicks is larger than 10, FCTSFN performs similarly to ESC and GSC, and it has better performance compared to the other methods. The proposed FCTSFN shows a larger advantage on the Pascal VOC 2012, Microsoft Coco and Berkeley datasets than on the Grabcut dataset. One possible reason is that the Grabcut dataset has a smaller number of images with more distinct foreground/background. Therefore, FCTSFN performs similarly to ESC and GSC on the Grabcut dataset given sufficient number of clicks. In summary, the proposed FCTSFN shows consistently improved performance on larger and more challenging datasets, while it still achieves stable and top performance on smaller and less challenging datasets. Tab.~\ref{tab_experiment_cmp} reports the mean number of clicks to achieve a certain IoU. It can be seen that the proposed FCTSFN needs the least number of clicks on all the datasets. Note that the best possible number of clicks to achieve a certain IoU is $1$. Therefore, the proposed FCTSFN achieves an improvement of ${\left( 5.81-4.58 \right)} / {\left( 5.81-1 \right)}\approx25.6\%$ towards the best possible performance with respect to SSFCN on the VOC 2012 dataset. This figure is $23.3\%$, $31.3\%$ and $26.6\%$ for the Microsoft Coco, Grabcut and Berkeley datasets, respectively. Fig.~\ref{fig_example_rst} shows some example results of different methods given the same user clicks. Fig.~\ref{fig_example_rst_auto_clicks} shows some example results of the proposed method on different objects in the test datasets, with automatically generated click sequences with up to 5 clicks.

\setlength{\tabcolsep}{4pt}
\begin{table*}[!t]
\begin{center}
\caption{Mean IoU at certain number of clicks for restricted comparisons (in percentage, best performance in bold)}
\label{tab_experiment_IoU_cmp}
%\resizebox{\textwidth}{!}{\begin{tabular}{cccccccc}
%\hline\noalign{\smallskip}
%Dataset & GC & GM & RW & ESC & GSC & SSFCN & FCTSFN \\
%\noalign{\smallskip}
%\hline
%\noalign{\smallskip}
%Pascal VOC 2012 (1 click) & 25.2 & 25.7 & 34.9 & 31.4 & 31.2 & 57.0 & {\bf 62.3} \\
%Microsoft Coco (1 click) & 18.1 & 20.3 & 32.3 & 26.9 & 26.6 & 38.6 & {\bf 42.5} \\
%Grabcut (1 click) & 49.6 & 38.6 & 39.8 & 48.2 & 51.2 & 76.8 & {\bf 77.7} \\
%Berkeley (1 click) & 43.8 & 33.6 & 34.6 & 46.7 & 46.7 & 68.8 & {\bf 74.5} \\
%\hline
%Pascal VOC 2012 (3 clicks) & 33.9 & 41.4 & 56.1 & 44.6 & 43.1 & 73.0 & {\bf 78.0} \\
%Microsoft Coco (3 clicks) & 25.7 & 35.2 & 52.0 & 38.5 & 36.7 & 55.1 & {\bf 61.2} \\
%Grabcut (3 clicks) & 66.2 & 58.4 & 56.1 & 74.4 & 74.6 & 83.2 & {\bf 87.9} \\
%Berkeley (3 clicks) & 51.6 & 47.1 & 55.0 & 58.4 & 60.1 & 81.0 & {\bf 84.8} \\
%\hline
%Pascal VOC 2012 (10 clicks) & 53.8 & 73.3 & 84.2 & 77.5 & 75.9 & 88.2 & {\bf 92.6} \\
%Microsoft Coco (10 clicks) & 39.8 & 65.4 & 77.4 & 69.9 & 66.6 & 72.8 & {\bf 81.5} \\
%Grabcut (10 clicks) & 84.3 & 82.5 & 86.8 & 91.8 & 91.0 & 91.0 & {\bf 94.7} \\
%Berkeley (10 clicks) & 70.8 & 76.9 & 80.6 & 83.0 & 83.8 & 89.4 & {\bf 92.6} \\
%\hline
%\end{tabular}}
\begin{tabular}{cccccccc}
\hline\noalign{\smallskip}
Dataset & GC & GM & RW & ESC & GSC & SSFCN & FCTSFN \\
\noalign{\smallskip}
\hline
\noalign{\smallskip}
Pascal VOC 2012 (1 click) & 25.2 & 25.7 & 34.9 & 31.4 & 31.2 & 57.0 & {\bf 62.3} \\
Microsoft Coco (1 click) & 18.1 & 20.3 & 32.3 & 26.9 & 26.6 & 38.6 & {\bf 42.5} \\
Grabcut (1 click) & 49.6 & 38.6 & 39.8 & 48.2 & 51.2 & 76.8 & {\bf 77.7} \\
Berkeley (1 click) & 43.8 & 33.6 & 34.6 & 46.7 & 46.7 & 68.8 & {\bf 74.5} \\
\hline
Pascal VOC 2012 (3 clicks) & 33.9 & 41.4 & 56.1 & 44.6 & 43.1 & 73.0 & {\bf 78.0} \\
Microsoft Coco (3 clicks) & 25.7 & 35.2 & 52.0 & 38.5 & 36.7 & 55.1 & {\bf 61.2} \\
Grabcut (3 clicks) & 66.2 & 58.4 & 56.1 & 74.4 & 74.6 & 83.2 & {\bf 87.9} \\
Berkeley (3 clicks) & 51.6 & 47.1 & 55.0 & 58.4 & 60.1 & 81.0 & {\bf 84.8} \\
\hline
Pascal VOC 2012 (10 clicks) & 53.8 & 73.3 & 84.2 & 77.5 & 75.9 & 88.2 & {\bf 92.6} \\
Microsoft Coco (10 clicks) & 39.8 & 65.4 & 77.4 & 69.9 & 66.6 & 72.8 & {\bf 81.5} \\
Grabcut (10 clicks) & 84.3 & 82.5 & 86.8 & 91.8 & 91.0 & 91.0 & {\bf 94.7} \\
Berkeley (10 clicks) & 70.8 & 76.9 & 80.6 & 83.0 & 83.8 & 89.4 & {\bf 92.6} \\
\hline
\end{tabular}
\end{center}
\end{table*}
\setlength{\tabcolsep}{1.4pt}

\setlength{\tabcolsep}{4pt}
\begin{table*}[!t]
\begin{center}
\caption{Mean number of clicks to achieve a certain IoU for resrticted comparisons (best performance in bold)}
\label{tab_experiment_cmp}
%\resizebox{\textwidth}{!}{\begin{tabular}{cccccccc}
%\hline\noalign{\smallskip}
%Dataset & GC & GM & RW & ESC & GSC & SSFCN & FCTSFN \\
%\noalign{\smallskip}
%\hline
%\noalign{\smallskip}
%Pascal VOC 2012 (85\% IoU) & 14.81 & 10.59 & 7.98 & 8.22 & 8.48 & 5.81 & {\bf 4.58} \\
%Microsoft Coco (85\% IoU) & 17.74 & 14.57 & 11.71 & 11.70 & 12.11 & 11.42 & {\bf 9.62} \\
%Grabcut (90\% IoU) & 9.70 & 9.26 & 10.28 & 5.84 & 5.02 & 5.02 & {\bf 3.76} \\
%Berkeley (90\% IoU) & 13.68 & 14.10 & 13.46 & 9.73 & 9.38 & 8.48 & {\bf 6.49} \\
%\hline
%\end{tabular}}
\begin{tabular}{cccccccc}
\hline\noalign{\smallskip}
Dataset & GC & GM & RW & ESC & GSC & SSFCN & FCTSFN \\
\noalign{\smallskip}
\hline
\noalign{\smallskip}
Pascal VOC 2012 (85\% IoU) & 14.81 & 10.59 & 7.98 & 8.22 & 8.48 & 5.81 & {\bf 4.58} \\
Microsoft Coco (85\% IoU) & 17.74 & 14.57 & 11.71 & 11.70 & 12.11 & 11.42 & {\bf 9.62} \\
Grabcut (90\% IoU) & 9.70 & 9.26 & 10.28 & 5.84 & 5.02 & 5.02 & {\bf 3.76} \\
Berkeley (90\% IoU) & 13.68 & 14.10 & 13.46 & 9.73 & 9.38 & 8.48 & {\bf 6.49} \\
\hline
\end{tabular}
\end{center}
\end{table*}
\setlength{\tabcolsep}{1.4pt}

{\bf Unrestricted comparison.} For the unrestricted comparison, we compare with the following methods: RIS-Net~\cite{RISNet_Liew_2017}, DEXTR~\cite{DEXTR_Maninis_2018} and latent diversity network (LDN)~\cite{LatentDensityInteractive_Li_2018}. We directly cite the number of clicks to achieve a certain IoU reported in these papers. Note that, as discussed above, these results are not directly comparable due to the differences in the experiment settings. For example, for the test data on Microsoft Coco dataset, different methods use different random sampling settings; these methods also adopt different training data and various training strategies; \textit{etc}. However, this comparison shows the performance of the proposed network among state-of-the-art methods with open choices for experimental settings.

\setlength{\tabcolsep}{4pt}
\begin{table*}[!t]
\begin{center}
\caption{Mean number of clicks to achieve a certain IoU for unresrticted comparisons (best performance in bold)}
\label{tab_experiment_cmp_unrestricted}
%\resizebox{\textwidth}{!}{\begin{tabular}{ccccc}
%\hline\noalign{\smallskip}
%Dataset & RIS-Net & DEXTR & LDN & FCTSFN \\
%\noalign{\smallskip}
%\hline
%\noalign{\smallskip}
%Pascal VOC 2012 (85\% IoU) & 5.12 & {\bf 4.00} & - & 4.58 \\
%Microsoft Coco (85\% IoU) & - & - & 7.89 & 9.62 \\
%Microsoft Coco seen categories (85\% IoU) & {\bf 5.98} & - & - & - \\
%Microsoft Coco unseen categories (85\% IoU) & 6.44 & - & - & - \\
%Grabcut (90\% IoU) & 5.00 & 4.00 & 4.79 & {\bf 3.76} \\
%Berkeley (90\% IoU) & {\bf 6.03} & - & - & 6.49 \\
%\hline
%\end{tabular}}
\begin{tabular}{ccccc}
\hline\noalign{\smallskip}
Dataset & RIS-Net & DEXTR & LDN & FCTSFN \\
\noalign{\smallskip}
\hline
\noalign{\smallskip}
Pascal VOC 2012 (85\% IoU) & 5.12 & {\bf 4.00} & - & 4.58 \\
Microsoft Coco (85\% IoU) & - & - & {\bf 7.89} & 9.62 \\
Microsoft Coco seen categories (85\% IoU) & 5.98 & - & - & - \\
Microsoft Coco unseen categories (85\% IoU) & 6.44 & - & - & - \\
Grabcut (90\% IoU) & 5.00 & 4.00 & 4.79 & {\bf 3.76} \\
Berkeley (90\% IoU) & {\bf 6.03} & - & - & 6.49 \\
\hline
\end{tabular}
\end{center}
\end{table*}
\setlength{\tabcolsep}{1.4pt}

Tab.~\ref{tab_experiment_cmp_unrestricted} reports the performance of all methods in unrestricted comparison. It can be seen that the proposed FCTSFN achieves competitive performance on Pascal VOC 2012, Grabcut and Berkeley datasets. Specifically, compared to RIS-Net, FCTSFN needs fewer clicks to achieve a certain IoU on Pascal VOC 2012 and Grabcut datasets; it needs 0.46 more clicks than RIS-Net to achieve $90\%$ IoU on Berkeley dataset. Compared to DEXTR, FCTSFN performs better on Grabcut dataset, and it needs $0.58$ more clicks to achieve a $85\%$ IoU on Pascal VOC 2012 dataset. Compared to LDN, FCTSFN achieves a better performance on Grabcut dataset. Note that, compared to the proposed FCTSFN, DEXTR achieves the reported performance with more training data (Pascal VOC 2012 + SBD~\cite{SBD_Hariharan_2011}), with an online hard example mining (OHEM)~\cite{OHEM_Shrivastava_2016} based training strategy and a more advanced base network (ResNet-101~\cite{ResidueNet_He_2016}); similarly, LDN achieve its reported performance with a larger training set (SBD) and a more advanced segmentation network (context aggregation network~\cite{CAN_Yu_2016,CAN_Chen_2017}). In contrast, the proposed FCTSFN achieves the performance in Tab.~\ref{tab_experiment_cmp_unrestricted} with less training data (Pascal VOC 2012 only), without hard mining on training data during the training process, and with a less advanced base network (VGG16).

On the other hand, it can be seen from Tab.~\ref{tab_experiment_cmp_unrestricted} that the proposed FCTSFN needs the most number of clicks to achieve an IoU of $85\%$ on Microsoft Coco dataset compared to RIS-Net and LDN. However, since each comparison algorithm adopts different random sampling settings, we cannot estimate the effect of such a sampling on the final performance. For example, our implementation of SSFCN on our Microsoft Coco test data (as shown in Tab.~\ref{tab_experiment_cmp}) results in lower performance than those reported in other implementations with other randomly sampled Microsoft Coco test data~\cite{RISNet_Liew_2017,LatentDensityInteractive_Li_2018}.

In this subsection, we showed that the proposed FCTSFN achieved improved performance compared to other comparison methods in restricted comparisons. In unrestricted comparisons, it also achieves competitive performance with less training data and with a less advanced based network compared to other methods.

\section{Conclusions}
\label{sec:conclusion}

In this paper, we proposed a novel fully convolutional two-stream fusion network (FCTSFN) for interactive image segmentation. The intention is to firstly use a two-stream late fusion network (TSLFN) to allow the user interactions to have more direct and higher impact on the segmentation results to achieve improved accuracy, then use a multi-scale refining network (MSRN) to refine the segmentation result at full resolution to address the resolution loss in TSLFN. We conduct comprehensive experimental analysis and comparisons on four benchmark datasets. The main findings are summarised as follows:
\begin{itemize}
    \item We experimentally validate that the two-stream structure in TSLFN allows the user interactions to have a higher impact on the segmentation results and it achieves improved performance compared to single-stream networks.
    \item We experimentally validate the significance of the interaction stream in the TSLFN: the TSLFN with the interaction stream performs better than the TSLFN without this stream. This means that the interaction stream in the proposed network successfully learn richer and more meaningful feature representations from individual user interaction data.
		\item We experimentally validate the design choice of the proposed TSLFN. We show that the proposed architecture achieves generally better performance compared to its variations, given the fixed base network.
    \item We experimentally validate that the foreground refining performed by the MSRN in the FCTSFN leads to a further improvement on the performance of the TSLFN.
    \item In restricted comparisons, the proposed FCTSFN achieves better performance compared to state-of-the-art methods.	
		\item In unrestricted comparisons, the proposed FCTSFN also achieves competitive performance with less training data and a less advanced base network, compared to state-of-the-art methods.	
\end{itemize}
Future works may focus on: (1) implementing the two-stream structure with more advanced base networks to achieve better performance; (2) conducting more experimental and theoretical analysis to gain a deeper insight into the two-stream network structure for interactive image segmentation.

\section*{Acknowledgments}

This work was supported by Ice Communication Limited and Innovate UK (project  KTP/10412).

\section*{References}

\bibliography{mybibfile}

\end{document}